\documentclass[letterpaper]{article} 
\usepackage{aaai25}  
\usepackage{times}  
\usepackage{helvet}  
\usepackage{courier}  
\usepackage[hyphens]{url}  
\usepackage{graphicx} 
\usepackage{natbib}  
\usepackage{caption} 
\usepackage{algorithm}
\usepackage{algorithmic}

\usepackage{subcaption}
\usepackage{enumitem}
\usepackage{colortbl}
\usepackage{xcolor}
\usepackage{amsmath}
\usepackage{booktabs}
\usepackage{tabularx}
\usepackage{rotating}

\newcommand\crule[3][black]{\textcolor{#1}{\rule{#2}{#3}}}
\definecolor{orange}{HTML}{ff7832}
\definecolor{pink}{HTML}{ffc0cb}
\definecolor{yellow}{HTML}{ffff00}
\definecolor{blue}{HTML}{0096f5}
\definecolor{cyan}{HTML}{00ffff}
\definecolor{dark-orange}{HTML}{ff7f00}
\definecolor{red2}{HTML}{ff0000}
\definecolor{light-yellow}{HTML}{fff096}
\definecolor{brown}{HTML}{873c00}
\definecolor{purple}{HTML}{a020f0}
\definecolor{dark-pink}{HTML}{ff00ff}
\definecolor{gray}{HTML}{8b8989}
\definecolor{dark-purple}{HTML}{4b004b}
\definecolor{light-green}{HTML}{96f050}
\definecolor{white}{HTML}{e6e6fa}
\definecolor{green2}{HTML}{00af00}

\newcommand{\PreserveBackslash}[1]{\let\temp=\\#1\let\\=\temp}
\newcolumntype{C}[1]{>{\PreserveBackslash\centering}p{#1}}
\newcolumntype{R}[1]{>{\PreserveBackslash\raggedleft}p{#1}}
\newcolumntype{L}[1]{>{\PreserveBackslash\raggedright}p{#1}}

\usepackage{newfloat}
\usepackage{listings}
\DeclareCaptionStyle{ruled}{labelfont=normalfont,labelsep=colon,strut=off} 
\floatstyle{ruled}
\newfloat{listing}{tb}{lst}{}
\floatname{listing}{Listing}
\title{A Spatiotemporal Approach to Tri-Perspective Representation for 3D Semantic Occupancy Prediction}
\author {
    Sathira Silva\equalcontrib\textsuperscript{\rm 1,\rm 2},
    Savindu Wannigama\equalcontrib\textsuperscript{\rm 1},
    Gihan Jayatilaka\textsuperscript{\rm 3},
    Muhammad Haris Khan\textsuperscript{\rm 2},
    Roshan Ragel\textsuperscript{\rm 1}
}
\affiliations {
    \textsuperscript{\rm 1}University of Peradeniya, Peradeniya 20400, Sri Lanka\\
    \{e17369,roshanr\}@eng.pdn.ac.lk\\
    \textsuperscript{\rm 2}Mohamed bin Zayed University of AI, Abu Dhabi, UAE\\
    \{sathira.silva,muhammad.haris\}@mbzuai.ac.ae\\
    \textsuperscript{\rm 3}University of Maryland, College Park, MD 20742, USA\\
    gihan@cs.umd.edu
}

\usepackage{bibentry}\makeatletter\newcommand{\@AAAI}{AAAI}\makeatother

\begin{document}

\maketitle

\begin{abstract}
Holistic understanding and reasoning in 3D scenes are crucial for the success of autonomous driving systems. The evolution of 3D semantic occupancy prediction as a pretraining task for autonomous driving and robotic applications captures finer 3D details compared to traditional 3D detection methods. Vision-based 3D semantic occupancy prediction is increasingly overlooked in favor of LiDAR-based approaches, which have shown superior performance in recent years. However, we present compelling evidence that there is still potential for enhancing vision-based methods. Existing approaches predominantly focus on spatial cues such as tri-perspective view (TPV) embeddings, often overlooking temporal cues. This study introduces S2TPVFormer, a spatiotemporal transformer architecture designed to predict temporally coherent 3D semantic occupancy. By introducing temporal cues through a novel Temporal Cross-View Hybrid Attention mechanism (TCVHA), we generate Spatiotemporal TPV (S2TPV) embeddings that enhance the prior process. Experimental evaluations on the nuScenes dataset demonstrate a significant \textbf{$+4.1$\%} of absolute gain in mean Intersection over Union (mIoU) for 3D semantic occupancy compared to baseline TPVFormer, validating the effectiveness of S2TPVFormer in advancing 3D scene perception.
\end{abstract}

\section{Introduction}

Accurate and comprehensive 3D scene understanding and reasoning are crucial for the advancement of robotic and autonomous driving systems \cite{bevformer, triperspective, surroundocc, monoscene}. This reasoning encompasses two essential dimensions: spatial reasoning and temporal reasoning. Vision-based approaches to 3D perception \cite{pointpillars, cylindrical, lmscnet, pointrcnn} present distinct advantages over LiDAR-based methods that rely on explicit depth measurements. Notably, vision-centric methods excel in identifying road elements, such as traffic lights and road signs, a task that proves challenging for LiDAR-based approaches. 


For an extended period, one of the most prominent 3D perception tasks has been 3D object detection \cite{Simonelli2019-hw, Wang2021-ks, bevformer, bevdet}, which are constrained by the limited expressiveness of their 3D bounding box outputs.

This limitation was overcome by generalizing the expression of one cuboid into a collection of smaller cubes (voxels) that can collectively approximate arbitrary shapes by the introduction of a vision-centric 3D semantic occupancy prediction (SOP) task \cite{monoscene}. 3D SOP aims to capture the intricate details of the surrounding scene, leveraging information derived from surrounding multi-camera images captured from different perspective views. TPVFormer \cite{triperspective} introduces a Tri-Perspective View (TPV) representation and Cross-View Hybrid Attention (CVHA) as a self-attention mechanism over the three planes, for compute-efficient 3D semantic occupancy prediction. 

Previous works \cite{bevformer, huang2022bevdet4d, li2021hdmapnet} have emphasized the importance of temporal fusion in 3D object detection. However, earlier approaches to 3D SOP \cite{triperspective, surroundocc, occ3d, occformer} frequently overlooked the benefits of leveraging temporal information. This is evidenced by TPVFormer relying solely on the spatially fused features of the current scene for semantic predictions. Spatial fusion is the process of fusing 2D-to-3D lifted features from multi-camera views into a unified spatial representation. Building on this foundation, we propose using Cross-View Hybrid Attention (CVHA) to exchange spatiotemporal information across tri-perspective views. This exchange can be achieved through temporal feature fusion using one of the following approaches:
\begin{enumerate}[label={(\arabic*)}]
    \item Fusion of historical data only through the Bird’s Eye View (BEV) plane.\label{item:bev-fusion}
    \item Fusion of historical data through all tri-perspective views.\label{item:tpv-fusion}
\end{enumerate}

Approach~\ref{item:bev-fusion} has been explored in the literature using BEV warping \cite{bevformer, huang2022bevdet4d, occnet, beverse}. To consider the possibility of implementing approach~\ref{item:tpv-fusion}, several important details must be taken into account. The pitch and roll of the ego vehicle are often ignored due to their insignificance. The more significant yaw axis is aligned parallel to the Front and Side planes in the TPV representation. Changes in the yaw angle cause shifts in the position of these planes resulting in occupying different slices of the ego-space at different timestamps. Therefore, warping is feasible only on the BEV plane, complicating the implementation of approach~\ref{item:tpv-fusion}. Additionally, BEV warping can lead to information loss. UniFusion \cite{unifusion}, a spatiotemporal transformer method for map segmentation, addresses this issue by introducing \textit{virtual views} for parallel and adaptive spatiotemporal fusion across all camera views and time steps.

To bridge the gaps identified in the 3D SOP literature, we propose \textbf{S2TPVFormer}; a unified spatiotemporal TPV encoder. We adopt TPV as the latent ego-space representation, harnessing the strengths of BEV and Voxel representations while maintaining computational efficiency. Our spatiotemporal transformer encoder produces temporally rich S2TPV embeddings, enabling the prediction of dense and temporally coherent 3D semantic occupancy through a lightweight MLP decoder. For the spatiotemporal fusion of multi-camera views into the TPV representation, we first transform historical camera views to the current time step using Virtual View Transformation (VVT) and then fuse the multi-camera features into the TPV representation for each time step. To facilitate the effective interaction of features across all time steps and TPV planes, we propose \textit{Temporal Cross-View Hybrid Attention (TCVHA)}. This mechanism allows features to interact not only within the same time step but also across different time steps, enhancing spatiotemporal context awareness and resulting in a unified spatiotemporal representation.

A summary of our main contributions is as follows:

\begin{itemize}
    \item We introduce S2TPVFormer, featuring a novel temporal fusion workflow for TPV representation, and demonstrate how CVHA facilitates the sharing of spatiotemporal information across the three planes.
    \item S2TPVFormer achieves significant improvements in the 3D SOP task on the nuScenes validation set, with a \textbf{+4.1\%} mIOU gain over the baseline TPVFormer, highlighting that vision-based 3D SOP still has considerable potential for improvement.
\end{itemize}
\section{Related Work}\label{related-work}

\subsubsection{Latent 3D Scene Representations:}\label{sec:3dscenerepresentation} 
\begin{figure}[ht!]
    \centering
    \begin{tabular}{*{3}{p{20mm}}}
        \multicolumn{3}{c}{\includegraphics[width=0.68\linewidth]{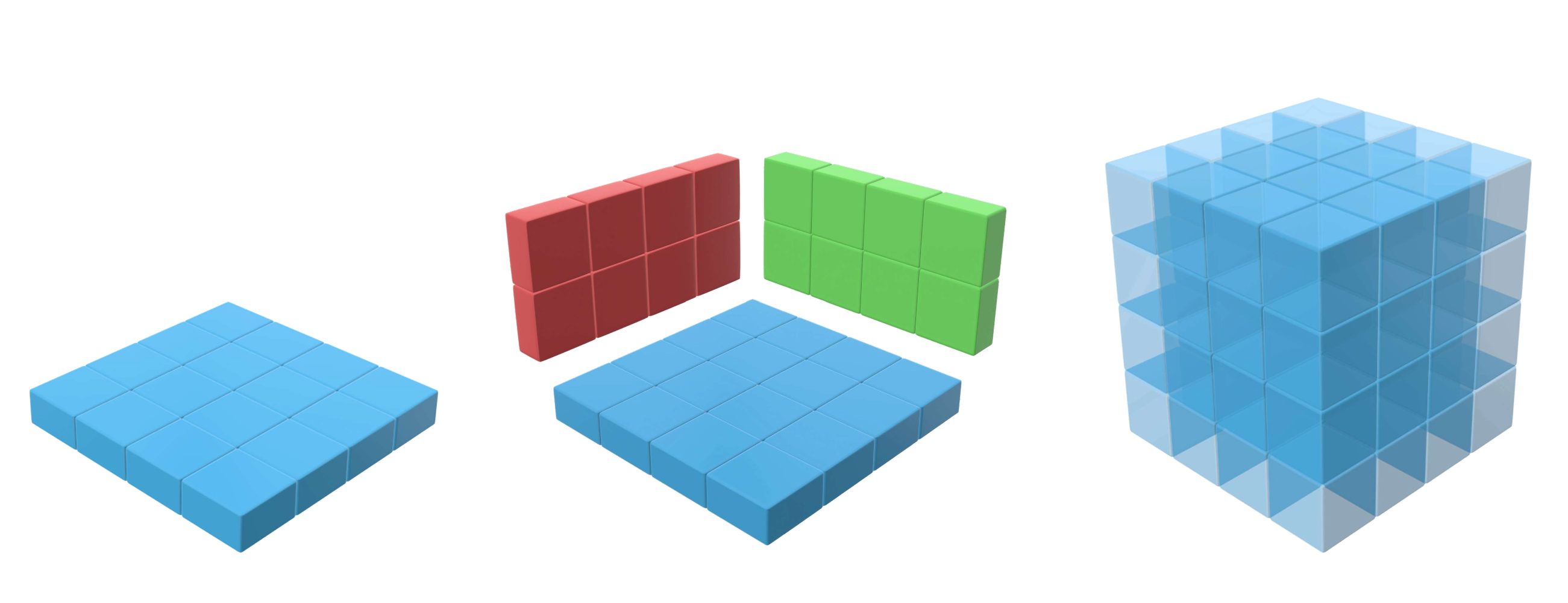}} \\
        \centering \scalebox{0.65}{$O(HW)$}     & \centering \scalebox{0.65}{$O(HW+DH+WD)$}     & \centering \scalebox{0.65}{$O(HWD)$}
    \end{tabular}
    \caption{\textbf{Comparison of BEV, TPV, and Voxel latent vector fields used to represent 3D scenes.}}
    \label{fig:3d_representationsl}
\end{figure}

The effectiveness of 3D scene understanding heavily relies on the representation of the 3D environment as illustrated in figure~\ref{fig:3d_representationsl}. Traditional approaches \cite{pseudolidar,imvoxelnet} involve dividing the 3D space into voxels and assigning each voxel a vector to denote its status, which is computationally expensive. Alternatively, BEV-based methods \cite{bevformer,bevdet,bevdepth,occbev,lss} perform remarkably well in tasks such as 3D object detection and map segmentation where height information is not significant. Some 3D SOP methods \cite{occnet} use BEV as the latent 3D scene embedding, but have to employ complex decoders to reconstruct the lost height information from BEV. TPVFormer \cite{triperspective} introduces a Tri-Perspective View (TPV) representation generalizing the BEV representation by incorporating two additional orthogonal planes.

\subsubsection{2D-3D View Transformation:} Transforming 2D perspective observations into 3D space latent embeddings can be considered an ill-posed problem due to the lack of depth information in 2D input images, but can be made feasible by incorporating a strong \textit{geometric prior}. Monocular single-camera approaches address this challenge by predicting explicit depth maps \cite{voxformer, lss}. For example, LSS \cite{lss} ``lifts'' each perspective view image individually into a frustum of features, then ``splats'' all frustums into a rasterized BEV grid. In contrast to LSS-based methods \cite{bevdepth,bevdet}, \textit{spatial fusion} is an alternative approach \cite{triperspective, bevformer, surroundocc, unifusion} which uses a spatial query-based transformer approach while leveraging camera parameters as a geometric prior to fuse spatial information from 2D perspective views into a unified latent representation for the ego-space. We adapt \textit{spatial fusion} since LSS-based view transformations tend to generate relatively sparse 3D representations.

\subsubsection{Temporal Reasoning:} Temporal reasoning holds equal importance to spatial reasoning in a cognitive perception system for identifying occluded objects and determining the motion state of entities. Spatial fusion provides a basis for temporal fusion. BEVFormer\cite{bevformer} recurrently fuses BEV features where the history features are warped to align with the ego-space of the current frame. The problem here is that since the warping occurs in ego BEV space which is pre-defined with bounded limits, some warped points are mapped outside the bounds of the original ego BEV space, leading to information loss. UniFusion \cite{unifusion} employs vanilla attention for attending to spatially mapped BEV features across all camera views and all time steps.  In this work we adapt this method of parallel spatiotemporal fusion.

\subsubsection{Vision-centric 3D Semantic Occupancy Prediction:}\label{sec:3dsop} The objective of \textit{3D SOP} is to intricately reconstruct the 3D environment surrounding an entity by incorporating detailed geometric information and semantic understanding. In the context of autonomous driving, 3D SOP serves as the academic alternative to occupancy networks \cite{mescheder2019occupancy}. 

MonoScene \cite{monoscene} is a pioneering work in vision-based 3D SOP, specifically focusing on Semantic Scene Completion (SSC). It introduces the first single-camera framework for SSC, enabling the reconstruction of outdoor scenes using RGB inputs alone. Building upon the foundation of MonoScene, TPVFormer \cite{triperspective}, the first multi-camera method for 3D SOP, introduces a tri-perspective view representation with a transformer-based TPV encoder. A more recent line of research \cite{occ3d, occnet, surroundocc} suggests that dense semantic occupancy predictions require dense labels and proposes pipelines for generating densified ground-truth voxel semantics. With our method, we demonstrate that leveraging temporal information provides an effective alternative to densifying supervision for achieving accurate SOP. 
\section{Methodology}\label{methodology}

\subsection{Overall Architecture} \label{sec:overall-architecture}

\begin{figure*}[ht!]
    \centering
    \includegraphics[width=.85\textwidth]{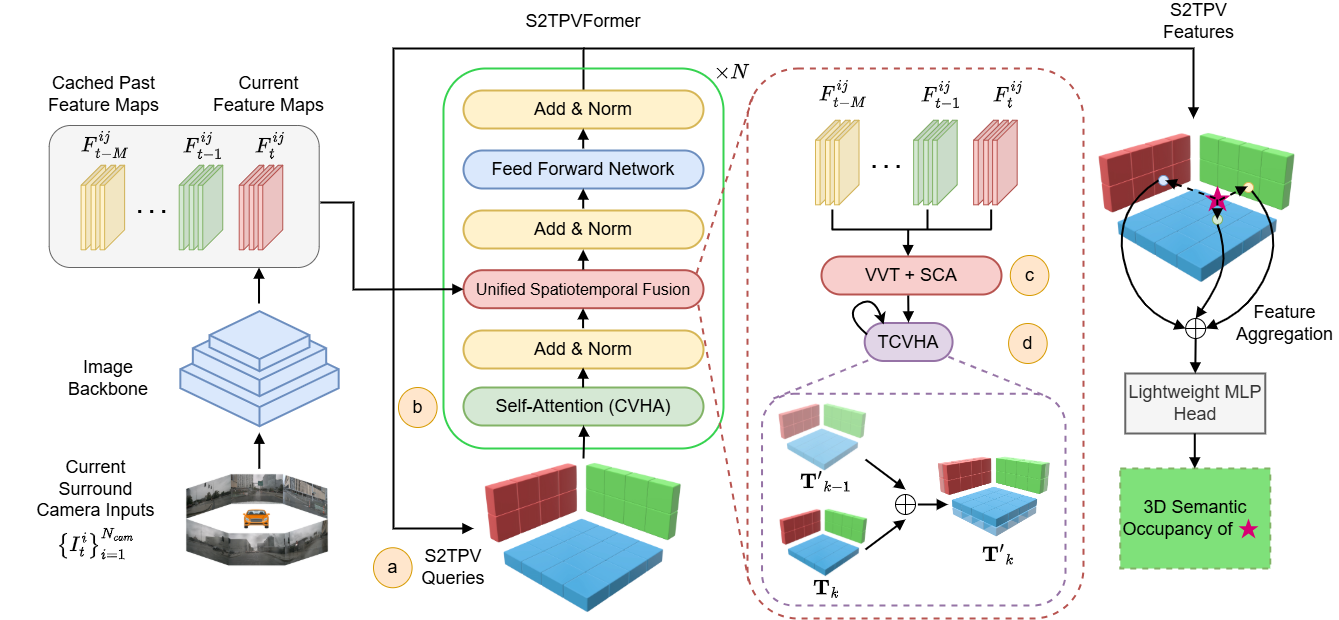}
    \caption{\textbf{The 3D SOP pipeline for the proposed S2TPVFormer architecture.} The S2TPVFormer encoder layers consist of four main components: (a) Three learnable grid-shaped parameters to learn spatiotemporal queries, (b) Self-Attention module, (c) Spatial Fusion (VVT + SCA) Module, and (d) Temporal Cross-View Hybrid Attention (TCVHA) Module. Both (c) and (d) are encapsulated as the \textit{Unified Spatiotemporal Fusion} Module in the block diagram.}
    \label{fig:S2TPVFormer}
\end{figure*}

Here we discuss our S2TPVFormer pipeline, which consists of four major modules, as illustrated in figure~\ref{fig:S2TPVFormer}. The 2D image backbone is detailed in the following section, and the spatiotemporal 2D-3D encoder is covered in the section after that. The remaining modules include a simple feature aggregator to generate voxel semantic occupancy features and a lightweight MLP head for predicting the semantic labels of individual voxels \cite{triperspective}.

\subsection{Image Backbone}\label{sec:image-backbone}

The image backbone consists of two networks; a feature extractor network and a neck module, which extracts multi-scale features for enhanced granularity. The image backbone network extracts multi-scale features from all the input surrounding multi-camera images simultaneously at a given timestep, providing the foundation for the S2TPV encoder. We employ a ResNet \cite{resnet} as the image feature extractor and an FPN \cite{lin2017feature} to produce multi-scale features. Given the $N_{cam}$ surround multi-camera images $\mathbf{I}_t$ at time step $t$, the image backbone is used to extract multi-level 2D perspective view features for each camera view. We denote these as $\mathbf{F}_t=\{\{F_t^{ij}\}_{j=1}^{N_{scale}}\}_{i=1}^{N_{cam}}$. Since the proposed pipeline is not limited to a specific image backbone, it can be replaced with any other feature extractor network such as ViT \cite{visiontransformer}, or SwinTransformer \cite{swin} along with any FPN variant such as BiFPN \cite{bifpn} or NAS-FPN \cite{ghiasi2019nasfpn}.

\subsection{S2TPV Encoder}\label{subsec:s2tpv-encoder}

During inference, S2TPVFormer caches $\mathbf{F}_t$ in a queue for each time step. These history feature maps, along with the current feature maps $\{\mathbf{F}_{t-k}^{ij}\}_{k=0}^{M}$, where $M$ is the total number of temporal fusion steps, are fed to the Unified Spatiotemporal Fusion module to fuse features across all camera views and time steps onto the S2TPV queries. Essentially, this module does the following, (1) Virtual View Transformation (VVT) to view camera features as if they were present in the current time step, followed by Spatial Cross Attention (SCA) to fuse virtual camera view features onto S2TPV queries for each time step, and (2) Fuse the virtual spatial TPV features across all time steps via TCVHA. The Temporal Cross-View Hybrid Attention (TCVHA), that we introduce extending CVHA, is realized via concatenating previous S2TPV features with current spatial TPV features as shown below the TCVHA module in figure~\ref{fig:S2TPVFormer}. A separate CVHA module is used to self-attend to S2TPV features to refine the queries and produce temporally coherent semantic occupancy embeddings. The S2TPV occupancy embeddings are finally aggregated and fed through a lightweight MLP head.

\subsubsection{Unified Spatiotemporal Fusion:} Since the spatial and temporal fusion in S2TPVFormer is parallel, history frames across the camera views has to be aligned with the current ego space. Given a past time frame, we use the VVT, as expressed in equations~\eqref{eq:vvt_rot} and ~\eqref{eq:vvt_trans}. In these equations, $R_i^{v,p}$ and $t_i^{v,p}$ represent the rotation and translation of the virtual view transformation for the $p^{th}$ time step. $R_i$ and $t_i$ denote the rotation and translation from the camera sensor to ego-space for the $i^{th}$ camera. $R_c$ and $t_c$ are the transformations from ego-space to global coordinates for the current time step, while $R_p$ and $t_p$ correspond to the ego-space to global coordinate transformation for the past time step. Together, $R_i^{v,p}$ and $t_i^{v,p}$ transform an ego-space point from a past time step to a virtual point in the current time step, as viewed from the perspective of the $i^{th}$ camera sensor. 

\begin{align}
    \label{eq:vvt_rot}
    R_i^{v,p} &= R_i^{-1}R_p^{-1}R_c \\
    \label{eq:vvt_trans}
    t_i^{v,p} &= R_i^{-1}R_p^{-1}t_c - R_i^{-1}R_p^{-1}t_p - R_i^{-1}t_i
\end{align}

We implement spatial fusion using 3D Deformable Attention \cite{bevformer} to reduce the computational burden of using vanilla attention. After the VVT, the virtual views are passed through the Spatial Cross-Attention (SCA) module to project them into the current ego TPV space. Taking advantage of the deformable attention mechanism, we implement VVT by employing the reference points defined in equation~\eqref{eq:cam_reference_pts}. For a S2TPV query $q_{h,w} \in Q^{HW}$ located at $(h,w)$, we uniformly sample $N_{ref}^{HW}$ reference points along the orthogonal direction of the plane as described in equation~\eqref{eq:ego_reference_pts} \cite{pointpillars}. These points are then transformed using the VVT and camera intrinsic parameters, which provide the geometry prior for the 2D-3D lifting, resulting in the final virtual image reference points. For the attention-based fusion, only the views where the projected reference point, $\mathbf{Ref}_{h,w}^{v,i}$, falls within the image bounds are considered. The SCA function that performs spatial fusion is described in equation~\eqref{eq:spatialcrossattention}.

\begin{align}
    \label{eq:ego_reference_pts}
    \mathbf{Ref}_{h,w}^{ego} &= \{(h,w,d_k)\}_{k=1}^{N_{ref}^{HW}} \\
    \label{eq:cam_reference_pts}
    \mathbf{Ref}_{h,w}^{v,i} &= K_iR_i^{v,p}\mathbf{Ref}_{h,w}^{ego} + t_i^{v,p} \\
    \label{eq:spatialcrossattention}
    \text{SCA}(q_{h,w}, \mathbf{F}_t) &= \\ \notag \frac{1}{|V_{hit}|}&\sum_{i\in V_{hit}}\text{3DDeformAttn}(q_{h,w},\mathbf{Ref}_{h,w}^{v,i},\mathbf{F}_t^i)
\end{align}

In equations \eqref{eq:ego_reference_pts} and \eqref{eq:cam_reference_pts}, $\mathbf{Ref}_{h,w}^{ego}$ represents the reference points generated in the ego-TPV space for each TPV plane, $\mathbf{Ref}_{h,w}^{v,i}$ denotes the $i^{th}$ virtual camera view reference points for 3D deformable attention, and $K_i$ is the camera intrinsic matrix of the $i^{th}$ camera sensor. In equation \eqref{eq:spatialcrossattention}, $V_{hit}$ denotes the set of hit camera views. Note that the above formulations consider only the $Q^{HW}$ query plane. The computations for the other two planes will follow the same approach. After spatial cross-attention, the resulting features are; 
\begin{align}
    \label{eq:resulting-features}
    \mathbf{T}_t = T_t^{HW}\cup T_t^{DH}\cup T_t^{WD}
\end{align}

\subsubsection{Temporal Cross-View Hybrid Attention:} Realizing the capability of CVHA in self-attending to S2TPV representation, we construct TCVHA, essentially for the queries to interact with history features. As depicted in figure~\ref{fig:S2TPVFormer}, for a given BEV query feature $q$ at a point $p=(h,w)$, it interacts with four types of feature points: (1) history points (temporal fusion), (2) self points, (3) front viewpoints, and (4) side viewpoints. Note that this diagram only illustrates interactions with BEV queries and does not show interactions with previous front and side view features. Given the spatially fused TPV features $\mathbf{T}_t$ for all the time steps in the history queue, the queries for the TCVHA are created iteratively, as described in equations~\eqref{eq:tcvha_queries}, ~\eqref{eq:deform_attn}, and ~\eqref{eq:intermediate_s2tpv}. Here, $q_{k, h, w}^{\prime} \in \mathbf{Q}_k^{\prime}$ represents the queries for the TCVHA operation at the $k^{th}$ iteration, and $\{\cdot\}$ denotes the concatenation operation. For the first iteration, the spatially fused TPV features from the last temporal fusion step, $\mathbf{T}_{t-M}$, are concatenated with themselves. The cross-view reference points, $\mathbf{Ref}_{h,w}^{cross}$, are generated in the same way as in \cite{triperspective}. Using these intermediate features as queries, TCVHA computes the temporally fused intermediate S2TPV features at $k^{th}$ iteration as expressed in equation~\eqref{eq:intermediate_s2tpv}. This is recursively repeated for $M$ number of temporal fusion steps until we get the final unified spatiotemporal features $\mathbf{T}_t^{\prime}$. Figure~\ref{fig:S2TPVFormer} illustrates the recurrent-style temporal fusion of TCVHA, shown below the TCVHA block.

\begin{align}
    \label{eq:tcvha_queries}
    \mathbf{Q}_k^{\prime} &= \{T_{k-1}^{\prime HW},T_k^{HW}\}\cup\{T_{k-1}^{\prime DH},T_k^{DH}\}\\ \notag &\cup\{T_{k-1}^{\prime WD},T_k^{WD}\}\\
    \label{eq:deform_attn}
    \text{TCVHA}(q_{k, h, w}^{\prime}) &= \text{DeformAttn}(q_{k, h, w}^{\prime}, \mathbf{Ref}_{h,w}^{cross}, \mathbf{T^{\prime}}_k) \\
    \label{eq:intermediate_s2tpv}
    \mathbf{T^{\prime}}_k &= \text{TCVHA}(q_{k, h, w}^{\prime})
\end{align}

\section{Experimental Setup and Implementation}

\subsection{Datasets and Evaluation Metrics}

We use nuScenes \cite{nuscenes}, which is a large-scale dataset for autonomous driving research, providing 1000 urban driving scenes with annotations for object detection. The dataset contains 28,130 training and 6,018 validation keyframes, captured at 20Hz. It incorporates data from monocular cameras, LiDAR, RADAR, and GPS, with 1.1 billion LiDAR points manually annotated for 32 classes.

For 3D perception tasks such as \textit{LiDAR Segmentation} and \textit{3D SOP}, mean Intersection over Union (mIoU) stands as a significant validation metric to measure model accuracy. A higher IoU signifies better overlap between predicted and ground truth bounding boxes or segmented regions, offering a precise assessment of localization accuracy. Complementing this, mIoU calculates the average IoU across multiple classes, providing an overall performance indicator.

\subsection{Training for 3D Semantic Occupancy Prediction and LiDAR Segmentation} \label{sec:spvsop}

\subsubsection{Ground Truth Labels for Supervision:} 

Models for both 3D SOP and LiDAR segmentation are trained with supervision from the training set of the nuScenes dataset. The nuScenes dataset comprises images, sparse LiDAR points from sweeps across the scene, and annotations of 16 semantic classes for each point. To train our models we divide the 3D space into a voxel grid and assign the semantic label of the LiDAR points that fall into a voxel as the semantic label of the voxel itself. A new semantic class, `empty', is assigned to all voxels without any LiDAR points falling onto them for training the 3D SOP model. This approach is in line with the standard practices proposed in various studies \cite{triperspective, surroundocc, occformer, occ3d}. It is important to mention that we do not generate super-resolution voxel semantic labels, as has been explored in some related work \cite{occ3d, surroundocc}.

\subsubsection{Loss Functions \& Training:}
We use two loss functions during training: (a) a Cross-entropy loss to improve voxel classification accuracy and (b) a Lovasz-softmax loss \cite{lovasz-softmax} to maximize the IoU score across classes. When training for \textbf{3D SOP}, we supervise voxel predictions with the Lovasz loss and LiDAR point predictions with Cross-Entropy loss. Conversely, when training for \textbf{LiDAR Segmentation}, the supervision is reversed, as suggested by the ablation study results of TPVFormer \cite{triperspective}. For both tasks we apply an equal-weighted summation to get the total loss. This approach helps the latent representation learn the discretization strategy inherent to the voxel space. We have also used several data augmentation techniques, including image scaling, color distortion, and Gridmask \cite{gridmask}.

\begin{table}[t]
\centering
\fontsize{9}{11}\selectfont 
\setlength{\tabcolsep}{10pt}
\begin{tabularx}{\linewidth}{l|cc}
\toprule
  Model Config &
  \begin{tabular}[c]{@{}c@{}}Embedding\\ Dimensionality\end{tabular} &
  Backbone \\ \midrule
S2TPVFormer (base)    & 256 & ResNet101  \\ 
S2TPVFormer (small)   & 128 & ResNet50   \\ \midrule
\end{tabularx}
\caption{\fontsize{10}{12}\selectfont \textbf{Model configurations used to run experiments}}
\label{table:configs}
\end{table}

\subsection{Implementation Details} \label{sec:experimental-design-implementation}

To highlight the encoder's efficiency, we use a lightweight MLP decoder composed of two linear layers with a Softplus activation layer in between. For different configurations in table~\ref{table:configs}, S2TPVFormer (base) employs a ResNet101-DCN \cite{Dai_2017_ICCV} initialized from an FCOS3D \cite{Wang2021-dp} checkpoint, while S2TPVFormer (small) employs a ResNet50 \cite{resnet} pre-trained on the ImageNet dataset \cite{imagenet-deng2009imagenet}. For both configurations, we set the input image resolution to 1600x900, TPV resolution to 100x100x8, and the number of transformer encoder layers to $N=3$ for all experiments, unless stated otherwise.
\section{Results and Analysis} \label{chap:results-analysis}

\subsection{Analysis of 3D Semantic Occupancy Prediction Results} \label{sec:analysis-sop-results}

\subsubsection{Quantitative Analysis:} \label{subsec:analysis-sop-results-quantitative}

\begin{table*}[]

\centering
\fontsize{9}{11}\selectfont 
\captionsetup{width=\textwidth}
\setlength{\tabcolsep}{3pt}
\begin{tabularx}{\textwidth}{l p{0.7cm} *{17}{>{\centering\arraybackslash}X}}
\toprule
  Method &
  \begin{sideways}mIoU (\%)\end{sideways} &
  \begin{sideways}\crule[orange]{0.2cm}{0.2cm} barrier\end{sideways}&
  \begin{sideways}\crule[pink]{0.2cm}{0.2cm} bicycle\end{sideways} &
  \begin{sideways}\crule[yellow]{0.2cm}{0.2cm} bus\end{sideways} &
  \begin{sideways}\crule[blue]{0.2cm}{0.2cm} car\end{sideways} &
  \begin{sideways}\crule[cyan]{0.2cm}{0.2cm} const. veh.\end{sideways} &
  \begin{sideways}\crule[dark-orange]{0.2cm}{0.2cm} motorcycle\end{sideways} &
  \begin{sideways}\crule[red2]{0.2cm}{0.2cm} pedestrian\end{sideways} &
  \begin{sideways}\crule[light-yellow]{0.2cm}{0.2cm} traffic cone\end{sideways} &
  \begin{sideways}\crule[brown]{0.2cm}{0.2cm} trailer\end{sideways} &
  \begin{sideways}\crule[purple]{0.2cm}{0.2cm} truck\end{sideways} &
  \begin{sideways}\crule[dark-pink]{0.2cm}{0.2cm} drive. surf.\end{sideways} &
  \begin{sideways}\crule[gray]{0.2cm}{0.2cm} other flat\end{sideways} &
  \begin{sideways}\crule[dark-purple]{0.2cm}{0.2cm} sidewalk\end{sideways} &
  \begin{sideways}\crule[light-green]{0.2cm}{0.2cm} terrain\end{sideways} &
  \begin{sideways}\crule[white]{0.2cm}{0.2cm} manmade\end{sideways} &
  \begin{sideways}\crule[green2]{0.2cm}{0.2cm} vegetation\end{sideways} \\ \midrule
  
TPVFormer &
  \underline{52.0} &
  \underline{59.6} &
  \textbf{26.3} &
  \underline{77.6} &
  \underline{74.1} &
  \underline{30.9} &
  \underline{47.5} &
  \textbf{41.8} &
  \underline{20.2} &
  \underline{44.9} &
  \underline{67.8} &
  \underline{86.3} &
  \underline{54.5} &
  \underline{55.5} &
  \underline{54.6} &
  \underline{47.5} &
  \underline{44.0} \\ 


  \midrule
  \rowcolor{violet!8}S2TPVFormer (Base) & 
  \textbf{56.1} & \textbf{60.1} &	16.5 &	\textbf{85.9} &	\textbf{74.3} &	\textbf{42.2} &	\textbf{51.5} &	\underline{37.0} &	\textbf{21.2} &	\textbf{49.4} &	\textbf{74.2} &	\textbf{86.4} &	\textbf{56.3} &	\textbf{57.9} &	\textbf{55.0} &	\textbf{65.4} &	\textbf{65.0}	 \\
  
  
  S2TPVFormer (Small) &  																
  43.4 &
  54.3 &
  \underline{17.2} &
  66.0 &
  69.5 &
  28.2 &
  22.8 &
  32.1 &
  15.1 &
  31.7 &
  59.6 &
  82.4 &
  49.9 &
  47.8 &
  47.4 &
  34.9 &
  36.0 \\ \bottomrule
\end{tabularx}
\caption{\fontsize{10}{12}\selectfont \textbf{3D Semantic Occupancy Prediction results on the nuScenes validation set.} It is fair to compare the results of TPVFormer and S2TPVFormer (Base) as our Base configuration is the same as the configuration TPVFormer has used for 3D SOP.}
\label{table:base-sop-results}
\end{table*}

The experimental results demonstrate that S2TPVFormer outperforms the TPVFormer baseline in 3D Semantic Occupancy Prediction (SOP). As shown in table~\ref{table:base-sop-results}, we achieve a \textbf{$4.1\%$} improvement over TPVFormer for SOP. This highlights the contribution of our temporal attention mechanism. It is also noteworthy that the IoU increases for fourteen out of the sixteen classes, demonstrating the robustness of the proposed methodology.

\subsubsection{Qualitative Analysis:} \label{subsec:analysis-sop-results-qualitative}

\begin{figure*}[t]
    \centering
    \includegraphics[width=1\textwidth]{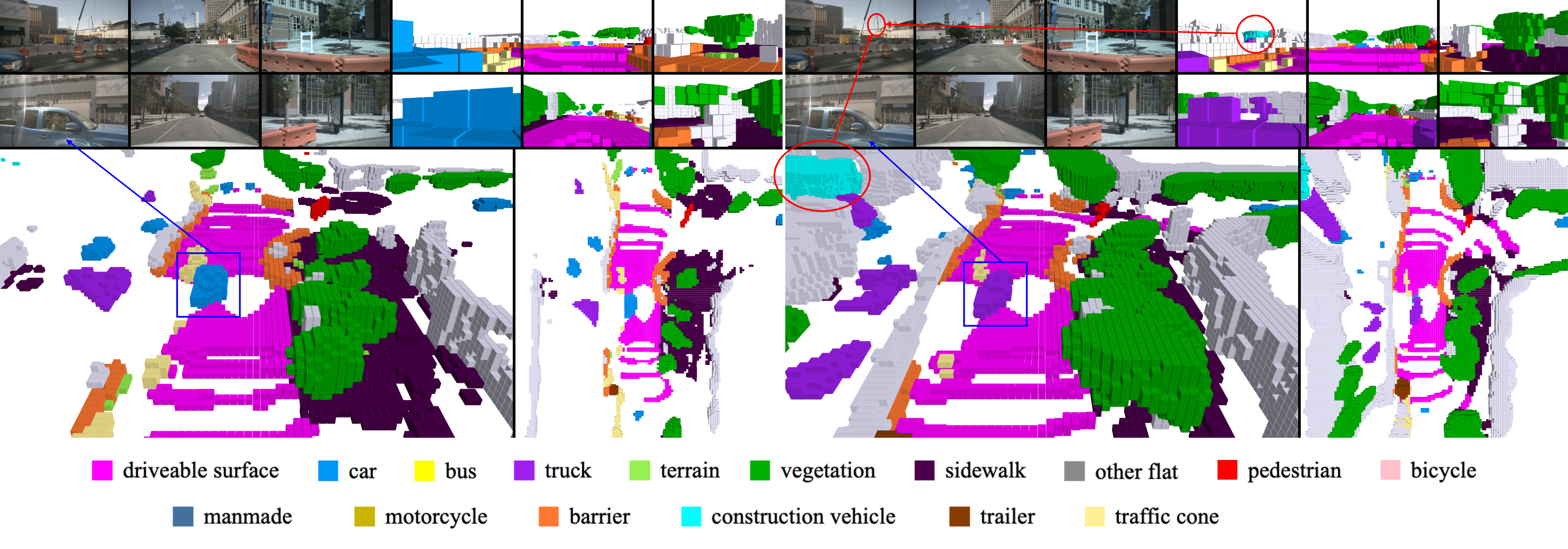}
    \caption{\textbf{Qualitative results on nuScenes validation set.} TPVFormer's \cite{triperspective} predictions are visualized on the left side, and S2TPVFormer's predictions are on the right side.}
    \label{fig:qualitative_1}
\end{figure*}

Figure~\ref{fig:qualitative_1} demonstrates the model's capability to predict 3D semantic occupancy around the ego vehicle. This figure presents six input camera images fed into the model, alongside eight representations of the semantic occupancy predictions made by the model for the same frame from the nuScene validation set. Through a comparative study with TPVFormer, we highlight the enhancements achieved through our novel temporal attention module.

Our analysis particularly focuses on two critical objects identified in the camera images for this frame; (a) a truck passing closely by the ego vehicle on its left, highlighted with a blue circle. TPVFormer misclassifies this truck as a car. We believe this is due to the truck’s proximity to the ego vehicle, which causes only the top half of the truck to be visible in the camera image, making it resemble a car. Conversely, S2TPVFormer accurately identifies it as a truck. We argue that this accuracy stems from the model's ability to integrate information from preceding frames, where the truck is captured in full from a distance. This allows the temporal fusion capability of S2TPVFormer to effectively utilize past frame data for accurate prediction. (b) A construction vehicle (more specifically a craine), highlighted with a red circle, visible in the distance in the front-left camera image. We contend that the model's access to temporally enriched image features enables S2TPVFormer to identify distant objects such as this.

Another notable observation from our analysis is that the predictions generated by S2TPVFormer are significantly denser than those of TPVFormer, even though both models are trained on the same sparse ground truth from the nuScenes dataset.

\subsection{Analysis of LiDAR Segmentation Results} \label{sec:analysis-lidarseg-results}

We test the performance of S2TPVFormer (base) for LiDAR segmentation to assess the generalization capabilities of our model, with a particular focus on the novel temporal attention module. We report the results of LiDAR segmentation on the nuScenes test and validation sets in tables~\ref{table:base-lidarseg-results-test} and \ref{table:base-lidarseg-results-val}, respectively. In table~\ref{table:base-lidarseg-results-test}, we present results for some of the best-performing methods in the nuScenes LiDAR segmentation challenge, including models that use both cameras and LiDAR as input modalities. Our model achieves promising results that are comparable with state-of-the-art methods in the literature.

\begin{table*}[hbt!]
\centering
\fontsize{9}{11}\selectfont 
\captionsetup{width=\textwidth}
\setlength{\tabcolsep}{3pt}
\begin{tabularx}{\textwidth}{l p{1cm} *{17}{>{\centering\arraybackslash}X}}
\toprule
Method &
\begin{tabular}[c]{@{}c@{}}Input\\ Modality\end{tabular} &
  \begin{sideways}mIoU ($\%$)\end{sideways} &
  \begin{sideways}\crule[orange]{0.2cm}{0.2cm} barrier\end{sideways}&
  \begin{sideways}\crule[pink]{0.2cm}{0.2cm} bicycle\end{sideways} &
  \begin{sideways}\crule[yellow]{0.2cm}{0.2cm} bus\end{sideways} &
  \begin{sideways}\crule[blue]{0.2cm}{0.2cm} car\end{sideways} &
  \begin{sideways}\crule[cyan]{0.2cm}{0.2cm} const. veh.\end{sideways} &
  \begin{sideways}\crule[dark-orange]{0.2cm}{0.2cm} motorcycle\end{sideways} &
  \begin{sideways}\crule[red2]{0.2cm}{0.2cm} pedestrian\end{sideways} &
  \begin{sideways}\crule[light-yellow]{0.2cm}{0.2cm} traffic cone\end{sideways} &
  \begin{sideways}\crule[brown]{0.2cm}{0.2cm} trailer\end{sideways} &
  \begin{sideways}\crule[purple]{0.2cm}{0.2cm} truck\end{sideways} &
  \begin{sideways}\crule[dark-pink]{0.2cm}{0.2cm} drive. surf.\end{sideways} &
  \begin{sideways}\crule[gray]{0.2cm}{0.2cm} other flat\end{sideways} &
  \begin{sideways}\crule[dark-purple]{0.2cm}{0.2cm} sidewalk\end{sideways} &
  \begin{sideways}\crule[light-green]{0.2cm}{0.2cm} terrain\end{sideways} &
  \begin{sideways}\crule[white]{0.2cm}{0.2cm} manmade\end{sideways} &
  \begin{sideways}\crule[green2]{0.2cm}{0.2cm} vegetation\end{sideways} \\ \midrule
 
MINet & LiDAR & 56.3 & 54.6 & 8.2 & 62.1 & 76.6 & 23.0 & 58.7 & 37.6 & 34.9 & 61.5 & 46.9 & 93.3 & 56.4 & 63.8 & 64.8 & 79.3 & 78.3 \\ 
LidarMultiNet & LiDAR & 81.4 & 80.4 & 48.4 & 94.3 & 90.0 & 71.5 & 87.2 & 85.2 & 80.4 & 86.9 & 74.8 & 97.8 & 67.3 & 80.7 & 76.5 & 92.1 & 89.6  \\

UniVision & LiDAR & 72.3 & 72.1 &	34.0 &	85.5 &	89.5 &	59.3 &	75.5 &	69.3 &	65.8 &	84.2 &	71.4 &	96.1 &	67.4 &	71.9 &	65 &	77.9 &	71.7 \\
PanoOcc & LiDAR & 71.4 & 82.5 &	32.3 &	88.1 &	83.7 &	46.1 &	76.5 &	67.6 &	53.6 &	82.9 &	69.5 &	96.0 &	66.3 &	72.3 &	66.3 &	80.5 &	77.3 \\
OccFormer & LiDAR & 70.8 & 72.8 &	29.9 &	87.9 &	85.6 &	57.1 &	74.9 &	63.2 &	53.5 &	83 &	67.6 &	94.8 &	61.9 &	70.0 &	66.0 &	84.0 &	80.5 \\
TPVFormer-Small\textsuperscript{\dag} & Camera & 59.2 & 65.6 & 15.7 & 75.1 & 80.0 & 45.8 & 43.1 & 44.3 & 26.8 & 72.8 & 55.9 & 92.3 & 53.7 & 61.0 & 59.2 & 79.7 & 75.6  \\
TPVFormer-Base\textsuperscript{\dag} & Camera & 69.4 & 74.0 & 27.5 & 86.3 & 85.5 & 60.7 & 68.0 & 62.1 & 49.1 & 81.9 & 68.4 & 94.1 & 59.5 & 66.5 & 63.5 & 83.8 & 79.9  \\ \midrule
 
S2TPVFormer (Base) & Camera & 60.4 &  61.2 &	18.2 &	80.6 &	78.1 &	55.2 &	57.6 &	41.5 &	26.4 &	76.1 &	61.3 &	89.8 &	49.4 &	56.6 &	58.0 &	79.3 &	76.4	 \\ 
\bottomrule

\end{tabularx}
\caption{\fontsize{10}{12}\selectfont \textbf{LiDAR Segmentation performance on the nuScenes test set.} \textsuperscript{\dag} represents that TPVFormer-Small and TPVFormer-Base are different from S2TPVFormer (small) and S2TPVFormer (base)}
\label{table:base-lidarseg-results-test}
\end{table*}

\begin{table*}[hbt!]
\centering
\fontsize{9}{11}\selectfont 
\setlength{\tabcolsep}{3pt}
\captionsetup{width=\textwidth}
\small
\begin{tabularx}{\textwidth}{l p{1cm} *{17}{>{\centering\arraybackslash}X}}
\toprule
Method &
\begin{tabular}[c]{@{}c@{}}Input\\ Modality\end{tabular} &
  \begin{sideways}mIoU\end{sideways} &
  \begin{sideways}\crule[orange]{0.2cm}{0.2cm} barrier\end{sideways}&
  \begin{sideways}\crule[pink]{0.2cm}{0.2cm} bicycle\end{sideways} &
  \begin{sideways}\crule[yellow]{0.2cm}{0.2cm} bus\end{sideways} &
  \begin{sideways}\crule[blue]{0.2cm}{0.2cm} car\end{sideways} &
  \begin{sideways}\crule[cyan]{0.2cm}{0.2cm} const. veh.\end{sideways} &
  \begin{sideways}\crule[dark-orange]{0.2cm}{0.2cm} motorcycle\end{sideways} &
  \begin{sideways}\crule[red2]{0.2cm}{0.2cm} pedestrian\end{sideways} &
  \begin{sideways}\crule[light-yellow]{0.2cm}{0.2cm} traffic cone\end{sideways} &
  \begin{sideways}\crule[brown]{0.2cm}{0.2cm} trailer\end{sideways} &
  \begin{sideways}\crule[purple]{0.2cm}{0.2cm} truck\end{sideways} &
  \begin{sideways}\crule[dark-pink]{0.2cm}{0.2cm} drive. surf.\end{sideways} &
  \begin{sideways}\crule[gray]{0.2cm}{0.2cm} other flat\end{sideways} &
  \begin{sideways}\crule[dark-purple]{0.2cm}{0.2cm} sidewalk\end{sideways} &
  \begin{sideways}\crule[light-green]{0.2cm}{0.2cm} terrain\end{sideways} &
  \begin{sideways}\crule[white]{0.2cm}{0.2cm} manmade\end{sideways} &
  \begin{sideways}\crule[green2]{0.2cm}{0.2cm} vegetation\end{sideways} \\ \midrule

BEVFormer & Camera & 56.2 & 54.0 & 22.8 & 76.7 & 74.0 & 45.8 & 53.1 & 44.5 & 24.7 & 54.7 & 65.5 & 88.5 & 58.1 & 50.5 & 52.8 & 71.0 & 63.0 \\
 
TPVFormer-Base\textsuperscript{\dag} & Camera & 68.9 & 70.0 & 40.9 & 93.7 & 85.6 & 49.8 & 68.4 & 59.7 & 38.2 & 65.3 & 83.0 & 93.3 & 64.4 & 64.3 & 64.5 & 81.6 & 79.3 \\

TPVFormer-Small\textsuperscript{\dag} & Camera & 59.3 & 64.9 & 27.0 & 83.0 & 82.8 & 38.3 & 27.4 & 44.9 & 24.0 & 55.4 & 73.6 & 91.7 & 60.7 & 59.8 & 61.1 & 78.2 & 76.5 \\ \midrule

S2TPVFormer (base) & Camera & 61.6 & 62.9 & 25.5 & 87.4 & 81.3 & 51.6 & 64.2 & 45.7 & 22.0 & 57.4 & 77.5 & 89.3 & 50.4 & 56.5 & 58.9 & 78.7 & 76.4 \\ 


\bottomrule

\end{tabularx}
\caption{\fontsize{10}{12}\selectfont \textbf{LiDAR Segmentation results on nuScenes validation set.} \textsuperscript{\dag} represents that TPVFormer-Small and TPVFormer-Base are different from S2TPVFormer (small) and S2TPVFormer (base)}
\label{table:base-lidarseg-results-val}
\end{table*}

\subsection{Ablation Study} \label{subsec:ablations}
We present two main ablation studies to investigate: (a) the range of temporal attention during inference, and (b) the dimensionality of the S2TPV embedding, in the context of 3D SOP.

\subsubsection{Range of Temporal Attention:} \label{subsubsec:ablation-temp-range} 

\begin{figure*}[!htp]
    \centering
    \begin{subfigure}[b]{0.3\textwidth}
        \centering
        \includegraphics[width=0.9\linewidth]{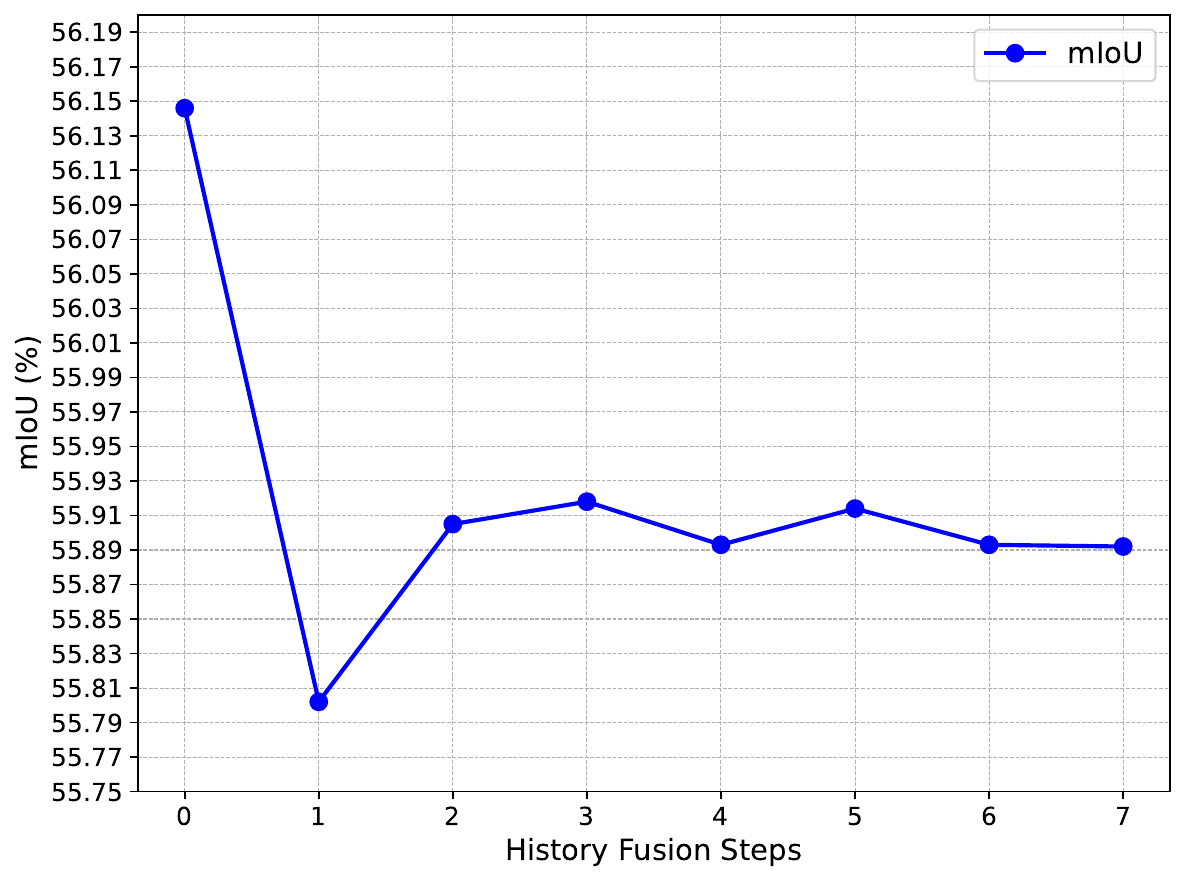}
        \caption{Mean IoU}
        \label{fig:ablation-temp-range1}  
    \end{subfigure}
    \hfill
    \begin{subfigure}[b]{0.3\textwidth}
        \centering
        \includegraphics[width=0.9\linewidth]{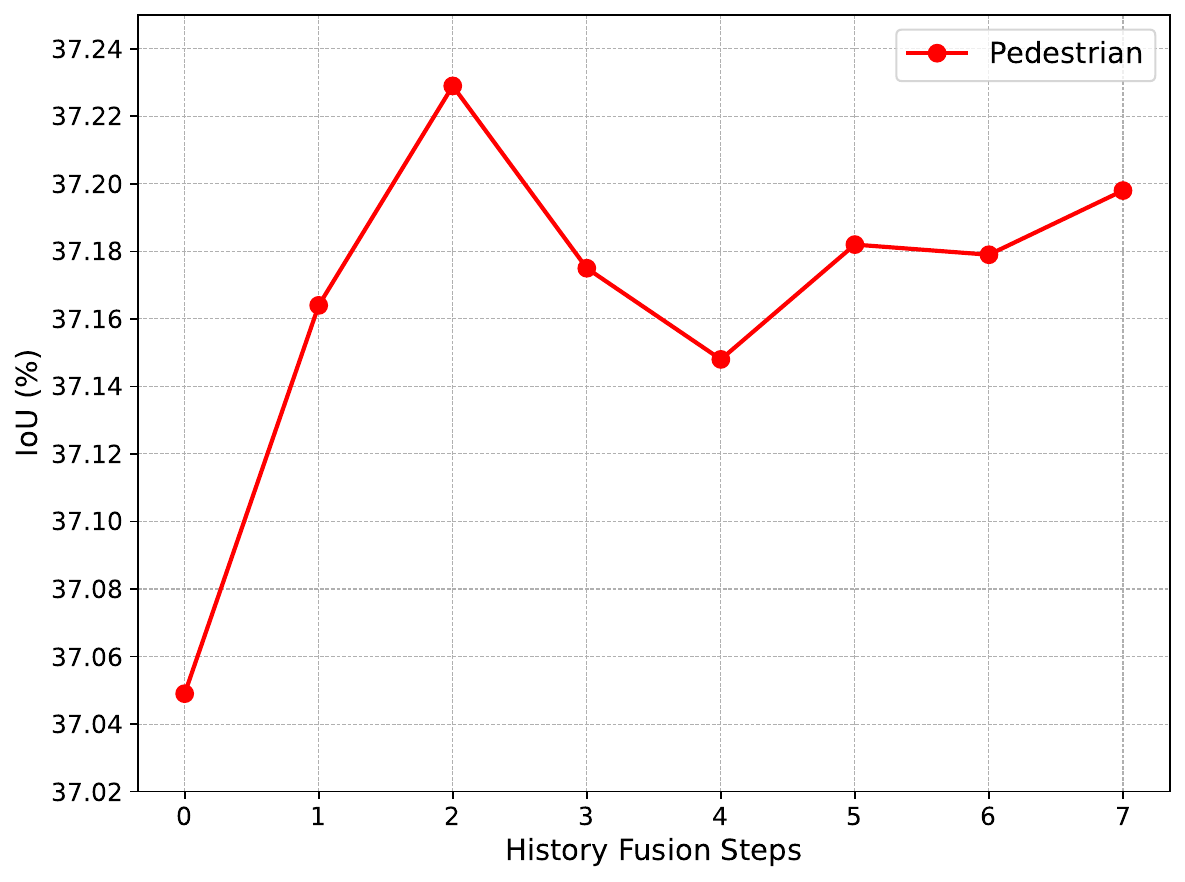}
        \caption{Pedestrian}
        \label{fig:ablation-temp-range2}  
    \end{subfigure}
    \hfill
    \begin{subfigure}[b]{0.3\textwidth}
        \centering
        \includegraphics[width=0.9\linewidth]{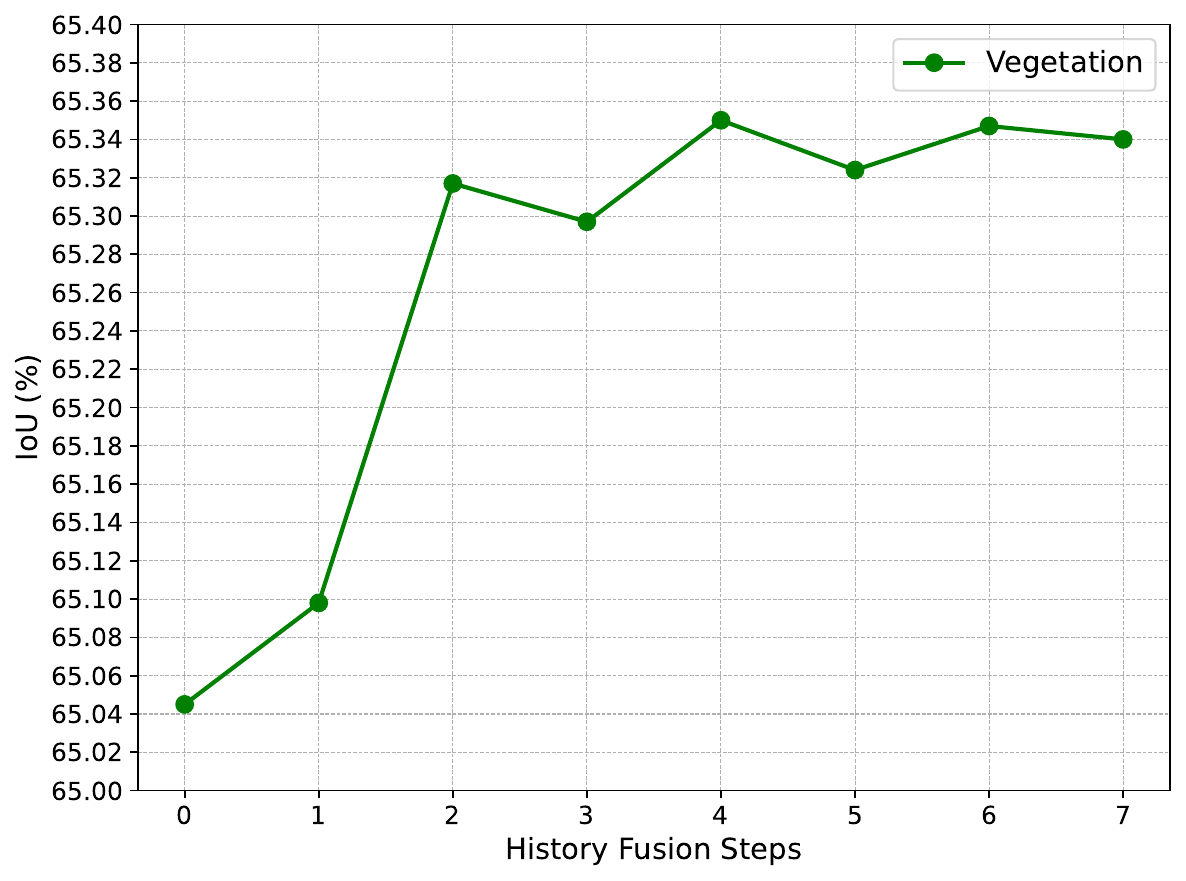}
        \caption{Vegetation}
        \label{fig:ablation-temp-range3}  
    \end{subfigure}
    \caption{Potential of long-range temporal fusion.}
    \label{fig:ablation-temp-range}
\end{figure*}

As discussed in section~\ref{subsec:s2tpv-encoder}, the training of our S2TPVFormer model is conducted using a single previous time frame for temporal attention. This study aims to examine the variation in performance of the model for 3D SOP as a function of varying extents of temporal attention. It is important to note that we change the history fusion steps only for inference.

As depicted in figure~\ref{fig:ablation-temp-range}, we present an analysis where the number of temporal history fusion steps is varied across eight different values, examining their impact on the IoU across two semantic classes as well as on the mean IoU. It is observed that the optimal number of history fusion steps necessary to achieve the most favorable outcomes differs among the semantic classes. This observation underscores the inherent potential for improving temporal fusion within our model, although it remains underexploited at the current juncture.

\begin{table}[h]
\centering
\fontsize{9}{11}\selectfont 
\setlength{\tabcolsep}{8pt}
\begin{tabular}{l|c}  
\toprule
Ablation & {mIoU (\%)} \\ \midrule
TPVFormer-Small\(^{*}\) & \bfseries 44.4 \\ 
S2TPVFormer (Small) & 43.4 \\ \midrule
TPVFormer & 52.0 \\
S2TPVFormer (Base) & \bfseries 55.0 \\
\bottomrule
\end{tabular}
\caption{\textbf{Summary of the ablation study on embedding dimensionality.} \(^{*}\) represents the reproduced results using our implementation of the TPVFormer's architecture.}
\label{table:ablation-small-sop-results}
\end{table}

\subsubsection{S2TPV Embedding Dimensionality:} \label{subsubsec:ablation-embed-dim}

For this study, we train S2TPVFormer and TPVFormer using the S2TPVFormer (small) configuration outlined in table~\ref{table:configs}. From the mIoU scores in table~\ref{table:ablation-small-sop-results}, we draw two important observations: \textbf{(a)} the mIoU scores of both TPVFormer and S2TPVFormer increase with the enhancement of embedding dimensionality, and \textbf{(b)} TPVFormer attains a higher mIoU than S2TPVFormer in the small configuration, even though the opposite is true for the base configuration. These observations lead us to conclude that \textbf{(a)} a higher embedding dimensionality is required to facilitate the TPV representation to learn and retain the additional information it receives via temporal attention, and \textbf{(b)} our model reveals promising scalability compared to TPVFormer.
\section{Conclusion}
Overshadowed by the increased performance of its LiDAR-based counterpart, the task of vision-based 3D Semantic Occupancy Prediction (3D SOP) has gradually lost traction within the academic community in the recent years. However, the vision-based approach still holds untapped potential for improvement. In this paper, we show one such improvement by introducing the novel approach of leveraging spatiotemporal information in the TPV representation to enhance the temporal coherence of 3D SOP. Our method specifically utilizes temporal attention to enhance the model's ability to comprehend and predict the 3D scene over time.

As the first to incorporate this method into the TPV representation, we demonstrate significant improvements in the accuracy of vision-based 3D SOP, reiterating its relevance despite the prominence of LiDAR methods. Our results show that incorporating temporal information can bridge some performance gaps between vision-based systems and their LiDAR counterparts. However, the full potential of long-range temporal information in these domains remains untapped. Future research should focus on further exploring our methodology, possibly focusing on the integration of dense semantic labels, to explore the complete capability of temporal attention in improving 3D scene understanding.

\section*{Acknowledgements}

We sincerely thank Honglu Zhou for providing Lambda credits, which enabled the execution of our experiments.


\bibliography{refs}

\begin{thebibliography}{38}
\providecommand{\natexlab}[1]{#1}

\bibitem[{Berman, Triki, and Blaschko(2018)}]{lovasz-softmax}
Berman, M.; Triki, A.~R.; and Blaschko, M.~B. 2018.
\newblock The lov{\'a}sz-softmax loss: A tractable surrogate for the optimization of the intersection-over-union measure in neural networks.
\newblock In \emph{Proceedings of the IEEE conference on computer vision and pattern recognition}, 4413--4421.

\bibitem[{Cao and de~Charette(2022)}]{monoscene}
Cao, A.-Q.; and de~Charette, R. 2022.
\newblock MonoScene: Monocular 3D Semantic Scene Completion.
\newblock In \emph{CVPR}.

\bibitem[{Chen et~al.(2020)Chen, Liu, Zhao, Wang, and Jia}]{gridmask}
Chen, P.; Liu, S.; Zhao, H.; Wang, X.; and Jia, J. 2020.
\newblock GridMask Data Augmentation.

\bibitem[{Dai et~al.(2017)Dai, Qi, Xiong, Li, Zhang, Hu, and Wei}]{Dai_2017_ICCV}
Dai, J.; Qi, H.; Xiong, Y.; Li, Y.; Zhang, G.; Hu, H.; and Wei, Y. 2017.
\newblock Deformable Convolutional Networks.
\newblock In \emph{Proceedings of the IEEE International Conference on Computer Vision (ICCV)}.

\bibitem[{Deng et~al.(2009)Deng, Dong, Socher, Li, Li, and Fei-Fei}]{imagenet-deng2009imagenet}
Deng, J.; Dong, W.; Socher, R.; Li, L.-J.; Li, K.; and Fei-Fei, L. 2009.
\newblock Imagenet: A large-scale hierarchical image database.
\newblock In \emph{2009 IEEE conference on computer vision and pattern recognition}, 248--255. Ieee.

\bibitem[{Dosovitskiy et~al.(2021)Dosovitskiy, Beyer, Kolesnikov, Weissenborn, Zhai, Unterthiner, Dehghani, Minderer, Heigold, Gelly, Uszkoreit, and Houlsby}]{visiontransformer}
Dosovitskiy, A.; Beyer, L.; Kolesnikov, A.; Weissenborn, D.; Zhai, X.; Unterthiner, T.; Dehghani, M.; Minderer, M.; Heigold, G.; Gelly, S.; Uszkoreit, J.; and Houlsby, N. 2021.
\newblock An Image is Worth 16x16 Words: Transformers for Image Recognition at Scale.
\newblock arXiv:2010.11929.

\bibitem[{Fong et~al.(2021)Fong, Mohan, Hurtado, Zhou, Caesar, Beijbom, and Valada}]{nuscenes}
Fong, W.~K.; Mohan, R.; Hurtado, J.~V.; Zhou, L.; Caesar, H.; Beijbom, O.; and Valada, A. 2021.
\newblock Panoptic nuScenes: A Large-Scale Benchmark for LiDAR Panoptic Segmentation and Tracking.
\newblock \emph{arXiv preprint arXiv:2109.03805}.

\bibitem[{Ghiasi et~al.(2019)Ghiasi, Lin, Pang, and Le}]{ghiasi2019nasfpn}
Ghiasi, G.; Lin, T.-Y.; Pang, R.; and Le, Q.~V. 2019.
\newblock NAS-FPN: Learning Scalable Feature Pyramid Architecture for Object Detection.
\newblock arXiv:1904.07392.

\bibitem[{He et~al.(2016)He, Zhang, Ren, and Sun}]{resnet}
He, K.; Zhang, X.; Ren, S.; and Sun, J. 2016.
\newblock Deep residual learning for image recognition.
\newblock In \emph{Proceedings of the IEEE conference on computer vision and pattern recognition}, 770--778.

\bibitem[{Huang and Huang(2022)}]{huang2022bevdet4d}
Huang, J.; and Huang, G. 2022.
\newblock BEVDet4D: Exploit Temporal Cues in Multi-camera 3D Object Detection.
\newblock \emph{arXiv preprint arXiv:2203.17054}.

\bibitem[{Huang et~al.(2021)Huang, Huang, Zhu, Ye, and Du}]{bevdet}
Huang, J.; Huang, G.; Zhu, Z.; Ye, Y.; and Du, D. 2021.
\newblock {BEVDet}: High-performance multi-camera {3D} object detection in {Bird-Eye-View}.

\bibitem[{Huang et~al.(2023)Huang, Zheng, Zhang, Zhou, and Lu}]{triperspective}
Huang, Y.; Zheng, W.; Zhang, Y.; Zhou, J.; and Lu, J. 2023.
\newblock Tri-Perspective View for Vision-Based 3D Semantic Occupancy Prediction.

\bibitem[{Lang et~al.(2019)Lang, Vora, Caesar, Zhou, Yang, and Beijbom}]{pointpillars}
Lang, A.~H.; Vora, S.; Caesar, H.; Zhou, L.; Yang, J.; and Beijbom, O. 2019.
\newblock PointPillars: Fast Encoders for Object Detection from Point Clouds.

\bibitem[{Li et~al.(2021)Li, Wang, Wang, and Zhao}]{li2021hdmapnet}
Li, Q.; Wang, Y.; Wang, Y.; and Zhao, H. 2021.
\newblock HDMapNet: An Online HD Map Construction and Evaluation Framework.

\bibitem[{Li et~al.(2022{\natexlab{a}})Li, Ge, Yu, Yang, Wang, Shi, Sun, and Li}]{bevdepth}
Li, Y.; Ge, Z.; Yu, G.; Yang, J.; Wang, Z.; Shi, Y.; Sun, J.; and Li, Z. 2022{\natexlab{a}}.
\newblock {BEVDepth}: Acquisition of reliable depth for multi-view {3D} object detection.

\bibitem[{Li et~al.(2023{\natexlab{a}})Li, Yu, Choy, Xiao, Alvarez, Fidler, Feng, and Anandkumar}]{voxformer}
Li, Y.; Yu, Z.; Choy, C.; Xiao, C.; Alvarez, J.~M.; Fidler, S.; Feng, C.; and Anandkumar, A. 2023{\natexlab{a}}.
\newblock VoxFormer: Sparse Voxel Transformer for Camera-based 3D Semantic Scene Completion.

\bibitem[{Li et~al.(2022{\natexlab{b}})Li, Wang, Li, Xie, Sima, Lu, Yu, and Dai}]{bevformer}
Li, Z.; Wang, W.; Li, H.; Xie, E.; Sima, C.; Lu, T.; Yu, Q.; and Dai, J. 2022{\natexlab{b}}.
\newblock {BEVFormer}: Learning bird's-eye-view representation from multi-camera images via spatiotemporal transformers.

\bibitem[{Li et~al.(2023{\natexlab{b}})Li, Yu, Wang, Anandkumar, Lu, and Alvarez}]{li2023fbbev}
Li, Z.; Yu, Z.; Wang, W.; Anandkumar, A.; Lu, T.; and Alvarez, J.~M. 2023{\natexlab{b}}.
\newblock Fb-bev: Bev representation from forward-backward view transformations.
\newblock In \emph{Proceedings of the IEEE/CVF International Conference on Computer Vision}, 6919--6928.

\bibitem[{Lin et~al.(2017)Lin, Dollár, Girshick, He, Hariharan, and Belongie}]{lin2017feature}
Lin, T.-Y.; Dollár, P.; Girshick, R.; He, K.; Hariharan, B.; and Belongie, S. 2017.
\newblock Feature Pyramid Networks for Object Detection.
\newblock arXiv:1612.03144.

\bibitem[{Liu et~al.(2021)Liu, Lin, Cao, Hu, Wei, Zhang, Lin, and Guo}]{swin}
Liu, Z.; Lin, Y.; Cao, Y.; Hu, H.; Wei, Y.; Zhang, Z.; Lin, S.; and Guo, B. 2021.
\newblock Swin Transformer: Hierarchical Vision Transformer using Shifted Windows.

\bibitem[{Mescheder et~al.(2019)Mescheder, Oechsle, Niemeyer, Nowozin, and Geiger}]{mescheder2019occupancy}
Mescheder, L.; Oechsle, M.; Niemeyer, M.; Nowozin, S.; and Geiger, A. 2019.
\newblock Occupancy Networks: Learning 3D Reconstruction in Function Space.

\bibitem[{Min et~al.(2023)Min, Xu, Li, Si, Xue, Jiang, Zhang, Li, Zhao, Xiao, Xu, Nie, and Dai}]{occbev}
Min, C.; Xu, X.; Li, F.; Si, S.; Xue, H.; Jiang, W.; Zhang, Z.; Li, J.; Zhao, D.; Xiao, L.; Xu, J.; Nie, Y.; and Dai, B. 2023.
\newblock Occ-BEV: Multi-Camera Unified Pre-training via 3D Scene Reconstruction.

\bibitem[{Philion and Fidler(2020)}]{lss}
Philion, J.; and Fidler, S. 2020.
\newblock Lift, splat, shoot: Encoding images from arbitrary camera rigs by implicitly unprojecting to {3D}.

\bibitem[{Qin et~al.(2022)Qin, Chen, Chen, Chen, and Li}]{unifusion}
Qin, Z.; Chen, J.; Chen, C.; Chen, X.; and Li, X. 2022.
\newblock {UniFusion}: Unified multi-view fusion transformer for spatial-temporal representation in bird's-eye-view.

\bibitem[{Rold{\~a}o, de~Charette, and Verroust-Blondet(2020)}]{lmscnet}
Rold{\~a}o, L.; de~Charette, R.; and Verroust-Blondet, A. 2020.
\newblock {LMSCNet}: Lightweight Multiscale {3D} Semantic Completion.

\bibitem[{Rukhovich, Vorontsova, and Konushin(2021)}]{imvoxelnet}
Rukhovich, D.; Vorontsova, A.; and Konushin, A. 2021.
\newblock ImVoxelNet: Image to Voxels Projection for Monocular and Multi-View General-Purpose 3D Object Detection.

\bibitem[{Shi, Wang, and Li(2018)}]{pointrcnn}
Shi, S.; Wang, X.; and Li, H. 2018.
\newblock {PointRCNN}: {3D} object proposal generation and detection from point cloud.

\bibitem[{Sima et~al.(2023)Sima, Tong, Wang, Chen, Wu, Deng, Gu, Lu, Luo, Lin, and Li}]{occnet}
Sima, C.; Tong, W.; Wang, T.; Chen, L.; Wu, S.; Deng, H.; Gu, Y.; Lu, L.; Luo, P.; Lin, D.; and Li, H. 2023.
\newblock Scene as Occupancy.

\bibitem[{Simonelli et~al.(2019)Simonelli, Bul{\`o}, Porzi, L{\'o}pez-Antequera, and Kontschieder}]{Simonelli2019-hw}
Simonelli, A.; Bul{\`o}, S. R.~R.; Porzi, L.; L{\'o}pez-Antequera, M.; and Kontschieder, P. 2019.
\newblock Disentangling monocular {3D} object detection.

\bibitem[{Tan, Pang, and Le(2020)}]{bifpn}
Tan, M.; Pang, R.; and Le, Q.~V. 2020.
\newblock EfficientDet: Scalable and Efficient Object Detection.
\newblock arXiv:1911.09070.

\bibitem[{Tian et~al.(2023)Tian, Jiang, Yun, Wang, Wang, and Zhao}]{occ3d}
Tian, X.; Jiang, T.; Yun, L.; Wang, Y.; Wang, Y.; and Zhao, H. 2023.
\newblock Occ3D: A Large-Scale 3D Occupancy Prediction Benchmark for Autonomous Driving.
\newblock \emph{arXiv preprint arXiv:2304.14365}.

\bibitem[{Wang et~al.(2021{\natexlab{a}})Wang, Zhu, Pang, and Lin}]{Wang2021-dp}
Wang, T.; Zhu, X.; Pang, J.; and Lin, D. 2021{\natexlab{a}}.
\newblock {FCOS3D}: Fully convolutional one-stage monocular {3D} object detection.

\bibitem[{Wang et~al.(2020)Wang, Chao, Garg, Hariharan, Campbell, and Weinberger}]{pseudolidar}
Wang, Y.; Chao, W.-L.; Garg, D.; Hariharan, B.; Campbell, M.; and Weinberger, K.~Q. 2020.
\newblock Pseudo-LiDAR from Visual Depth Estimation: Bridging the Gap in 3D Object Detection for Autonomous Driving.

\bibitem[{Wang et~al.(2021{\natexlab{b}})Wang, Guizilini, Zhang, Wang, Zhao, and Solomon}]{Wang2021-ks}
Wang, Y.; Guizilini, V.; Zhang, T.; Wang, Y.; Zhao, H.; and Solomon, J. 2021{\natexlab{b}}.
\newblock {DETR3D}: {3D} Object Detection from Multi-view Images via {3D-to-2D} Queries.

\bibitem[{Wei et~al.(2023)Wei, Zhao, Zheng, Zhu, Zhou, and Lu}]{surroundocc}
Wei, Y.; Zhao, L.; Zheng, W.; Zhu, Z.; Zhou, J.; and Lu, J. 2023.
\newblock SurroundOcc: Multi-Camera 3D Occupancy Prediction for Autonomous Driving.

\bibitem[{Zhang, Zhu, and Du(2023)}]{occformer}
Zhang, Y.; Zhu, Z.; and Du, D. 2023.
\newblock {OccFormer}: Dual-path transformer for vision-based {3D} semantic occupancy prediction.

\bibitem[{Zhang et~al.(2022)Zhang, Zhu, Zheng, Huang, Huang, Zhou, and Lu}]{beverse}
Zhang, Y.; Zhu, Z.; Zheng, W.; Huang, J.; Huang, G.; Zhou, J.; and Lu, J. 2022.
\newblock BEVerse: Unified Perception and Prediction in Birds-Eye-View for Vision-Centric Autonomous Driving.

\bibitem[{Zhu et~al.(2021)Zhu, Zhou, Wang, Hong, Li, Ma, Li, Yang, and Lin}]{cylindrical}
Zhu, X.; Zhou, H.; Wang, T.; Hong, F.; Li, W.; Ma, Y.; Li, H.; Yang, R.; and Lin, D. 2021.
\newblock Cylindrical and asymmetrical {3D} convolution networks for {LiDAR-based} perception.

\end{thebibliography}

\section{Supplementary Materials: A Spatiotemporal Approach to Tri-Perspective Representation for 3D Semantic Occupancy Prediction}

\begin{table*}[hbt!]

\centering
\caption{\textbf{3D Semantic Occupancy Prediction results on the nuScenes validation set.}}
\scalebox{.75}{
\begin{tabular}{@{}L{3.7cm}|C{0.65cm}|C{0.65cm}C{0.65cm}C{0.65cm}C{0.65cm}C{0.65cm}C{0.65cm}C{0.65cm}C{0.65cm}C{0.65cm}C{0.65cm}C{0.65cm}C{0.65cm}C{0.65cm}C{0.65cm}C{0.65cm}C{0.65cm}@{}}
\toprule
\multicolumn{1}{c|}{Method} &
  \begin{sideways}mIoU (\%)\end{sideways} &
  \begin{sideways}\crule[orange]{0.2cm}{0.2cm} barrier\end{sideways}&
  \begin{sideways}\crule[pink]{0.2cm}{0.2cm} bicycle\end{sideways} &
  \begin{sideways}\crule[yellow]{0.2cm}{0.2cm} bus\end{sideways} &
  \begin{sideways}\crule[blue]{0.2cm}{0.2cm} car\end{sideways} &
  \begin{sideways}\crule[cyan]{0.2cm}{0.2cm} const. veh.\end{sideways} &
  \begin{sideways}\crule[dark-orange]{0.2cm}{0.2cm} motorcycle\end{sideways} &
  \begin{sideways}\crule[red2]{0.2cm}{0.2cm} pedestrian\end{sideways} &
  \begin{sideways}\crule[light-yellow]{0.2cm}{0.2cm} traffic cone\end{sideways} &
  \begin{sideways}\crule[brown]{0.2cm}{0.2cm} trailer\end{sideways} &
  \begin{sideways}\crule[purple]{0.2cm}{0.2cm} truck\end{sideways} &
  \begin{sideways}\crule[dark-pink]{0.2cm}{0.2cm} drive. surf.\end{sideways} &
  \begin{sideways}\crule[gray]{0.2cm}{0.2cm} other flat\end{sideways} &
  \begin{sideways}\crule[dark-purple]{0.2cm}{0.2cm} sidewalk\end{sideways} &
  \begin{sideways}\crule[light-green]{0.2cm}{0.2cm} terrain\end{sideways} &
  \begin{sideways}\crule[white]{0.2cm}{0.2cm} manmade\end{sideways} &
  \begin{sideways}\crule[green2]{0.2cm}{0.2cm} vegetation\end{sideways} \\ \midrule

S2TPVFormer-W (Small) & 
  41.6 & 54.1 &	12.7 &	59.1 &	56.7 &	24.5 &	26.3 &	27.1 &	11.1 &	32.6 &	51.5 &	84 &	50.1 &	49.5 &	51.4 &	38.3 &	36.3	 \\
S2TPVFormer-W (Base) & 
  0.0 & 0.0 &	0.0 &	0.0 &	0.0 &	0.0 &	0.0 &	0.0 &	0.0 &	0.0 &	0.0 &	0.0 &	0.0 &	0.0 &	0.0 &	0.0 &	0.0	 \\
  
 \bottomrule
\end{tabular}
}
\label{table:base-sop-results}
\end{table*}

\begin{table*}[t]

\centering
\caption{\textbf{LiDAR Segmentation results on the nuScenes validation set.}}
\scalebox{.75}{
\begin{tabular}{@{}L{3.7cm}|C{0.65cm}|C{0.65cm}C{0.65cm}C{0.65cm}C{0.65cm}C{0.65cm}C{0.65cm}C{0.65cm}C{0.65cm}C{0.65cm}C{0.65cm}C{0.65cm}C{0.65cm}C{0.65cm}C{0.65cm}C{0.65cm}C{0.65cm}@{}}
\toprule
\multicolumn{1}{c|}{Method} &
  \begin{sideways}mIoU (\%)\end{sideways} &
  \begin{sideways}\crule[orange]{0.2cm}{0.2cm} barrier\end{sideways}&
  \begin{sideways}\crule[pink]{0.2cm}{0.2cm} bicycle\end{sideways} &
  \begin{sideways}\crule[yellow]{0.2cm}{0.2cm} bus\end{sideways} &
  \begin{sideways}\crule[blue]{0.2cm}{0.2cm} car\end{sideways} &
  \begin{sideways}\crule[cyan]{0.2cm}{0.2cm} const. veh.\end{sideways} &
  \begin{sideways}\crule[dark-orange]{0.2cm}{0.2cm} motorcycle\end{sideways} &
  \begin{sideways}\crule[red2]{0.2cm}{0.2cm} pedestrian\end{sideways} &
  \begin{sideways}\crule[light-yellow]{0.2cm}{0.2cm} traffic cone\end{sideways} &
  \begin{sideways}\crule[brown]{0.2cm}{0.2cm} trailer\end{sideways} &
  \begin{sideways}\crule[purple]{0.2cm}{0.2cm} truck\end{sideways} &
  \begin{sideways}\crule[dark-pink]{0.2cm}{0.2cm} drive. surf.\end{sideways} &
  \begin{sideways}\crule[gray]{0.2cm}{0.2cm} other flat\end{sideways} &
  \begin{sideways}\crule[dark-purple]{0.2cm}{0.2cm} sidewalk\end{sideways} &
  \begin{sideways}\crule[light-green]{0.2cm}{0.2cm} terrain\end{sideways} &
  \begin{sideways}\crule[white]{0.2cm}{0.2cm} manmade\end{sideways} &
  \begin{sideways}\crule[green2]{0.2cm}{0.2cm} vegetation\end{sideways} \\ \midrule

  
S2TPVFormer-W (Small) & 
  0.0 & 0.0 &	0.0 &	0.0 &	0.0 &	0.0 &	0.0 &	0.0 &	0.0 &	0.0 &	0.0 &	0.0 &	0.0 &	0.0 &	0.0 &	0.0 &	0.0	 \\
S2TPVFormer-W (Base) & 
  0.0 & 0.0 &	0.0 &	0.0 &	0.0 &	0.0 &	0.0 &	0.0 &	0.0 &	0.0 &	0.0 &	0.0 &	0.0 &	0.0 &	0.0 &	0.0 &	0.0	 \\
  
 \bottomrule
\end{tabular}
}
\label{table:base-sop-results}
\end{table*}

\subsection{Virtual View Transformation (VVT)}

We employ the concept of \textit{virtual camera views} introduced in \cite{unifusion} for our view transformation process, effectively mitigating the lossy temporal fusion associated with warp-based methods \cite{bevformer, occnet}. In this section, we offer an intuitive explanation of the underlying mechanics of VVT. Additionally, we provide an animated video that visually deconstructs the process for enhanced comprehension.

Three types of coordinate systems are involved in the view transformation process: (1) global/world coordinate system, (2) ego coordinate system, and (3) the $i^{th}$ camera sensor's coordinate system. The ego coordinate system is relative to the vehicle, where the origin is the vehicle's current position, and the camera sensor's coordinate system represents the world from the perspective of the $i^{th}$ camera's pose. As described in Sec 3.3 of our main paper, three transformation matrices are involved in the transformation. Among these, $R_c$ and $R_p$ have the same domain and codomain (ego to world), where the former is at the current time step and the latter at a past time step.

To map a point in the ego coordinate system to a point in a virtual camera view, the following steps are undertaken. Given a point $\mathbf{x}$ in the ego coordinate system, where $\mathbf{x}=(x,y,z,1)^T$, we first transform it to the world coordinate system as if it was initially viewed by the ego-vehicle at the current time step. $R_p^{-1}$ plays a major role in moving through space and time. Relative to the $R_c\mathbf{x}$ point, $R_p^{-1}$ moves the ego-vehicle back to where it was at the past time step, and views the point from there resulting in the virtual ego-space point $R_p^{-1}R_c\mathbf{x}$. Finally, it is viewed from the perspective of the $i^{th}$ camera sensor of interest through $R_i^{-1}$.

\subsection{Exploring the Confusion Matrix} \label{sec:conf_mat}

\begin{figure}[ht!]
    \centering
    \includegraphics[width=\linewidth]{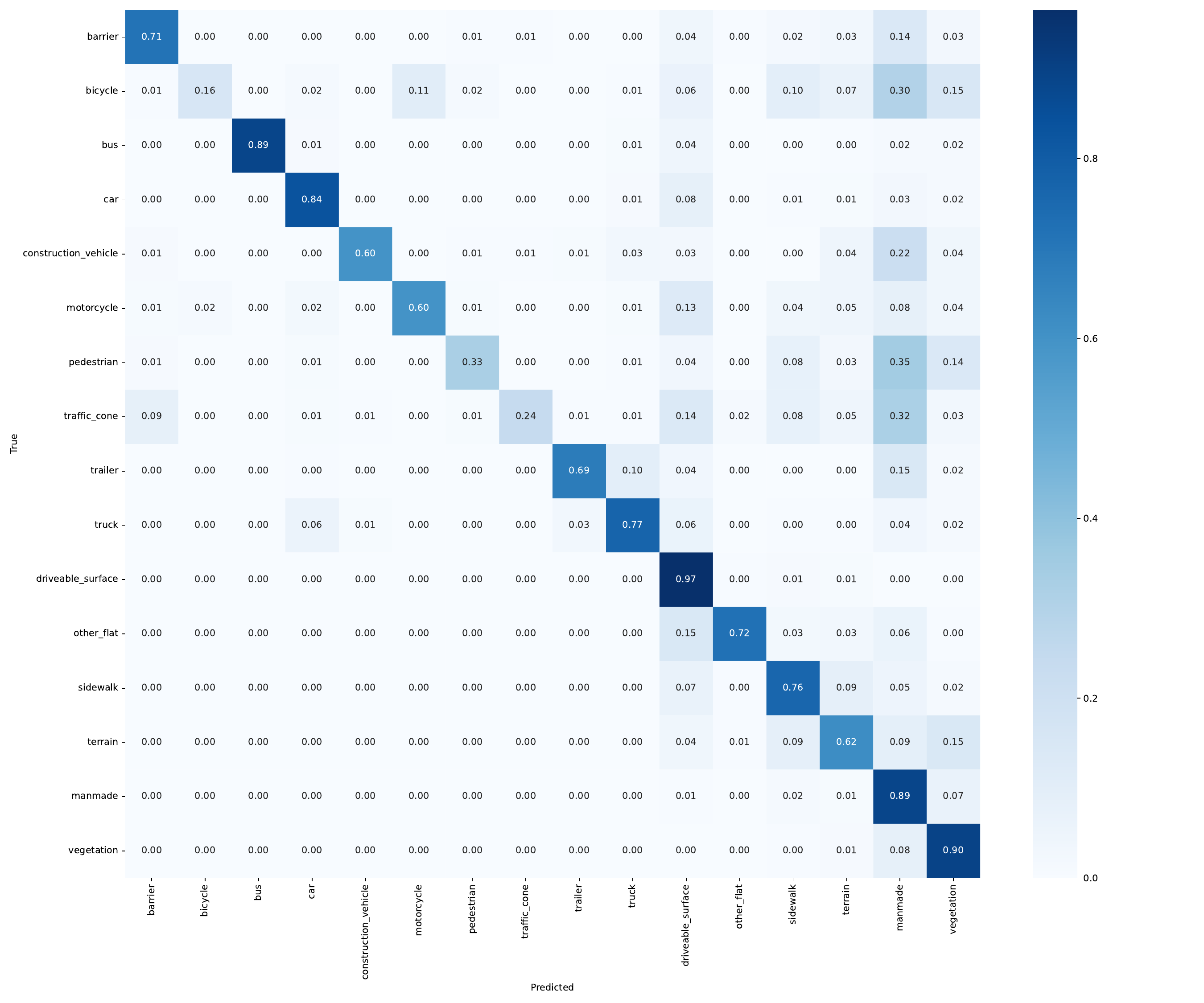}
    \caption{This figure presents the confusion matrix of the S2TPVFormer-U (base) model's predictions. It is important to note that this confusion matrix corresponds to the same predictions analyzed in our paper, where we detail the per-class IoUs and the mean IoU for 3D Semantic Occupancy Prediction (SOP) on the nuScenes validation dataset.}
    \label{fig:conf_mat}
\end{figure}

Figure \ref{fig:conf_mat} shows the confusion matrix for the same 3D SOP predictions on the nuScenes validation set analyzed in our paper. The model excels in identifying certain categories, achieving high precision -— true positive as a ratio of all positives —- in recognizing drivable surfaces (97\%), buses (89\%), and cars (83\%), as well as distinguishing environmental features like manmade objects (88\%) and vegetation (90\%), demonstrating its reliability in these areas. The confusion matrix also indicates the model's difficulty in correctly identifying certain classes, notably misclassifying barriers and bicycles as `manmade' or `vegetation', likely due to their similarity or presence in diverse environments. The high performance on large vehicles and environmental features indicates that the model's architecture is suitable for capturing and recognizing larger and more distinct shapes. This is further discussed in Sec. \ref{sec:big_small-objects}.

\subsection{S2TPVFormer Performance on Different Sizes of Objects}\label{sec:big_small-objects}

Our analysis reveals a consistent trend wherein our model demonstrates suboptimal performance in accurately predicting semantic occupancy for smaller, dynamic objects. Within the scope of the sixteen semantic classes, the three classes representing the smallest objects are bicycles, pedestrians, and traffic cones. Notably, among these, the traffic cone represents a static object, in contrast to the bicycles and pedestrians, which are inherently dynamic. Despite S2TPVFormer-U (base) surpassing TPVFormer across 14 out of 16 classes in terms of per-class Intersection over Union (IoU), for bicycles and pedestrians TPVFormer achieves better results. This outcome suggests a potential limitation of our model in retaining focus on smaller dynamic objects through the process of learning features from historical frames via temporal attention, as it appears to be focusing more on larger (static) objects.

\begin{figure}[h]
    \centering
    \includegraphics[width=1\linewidth]{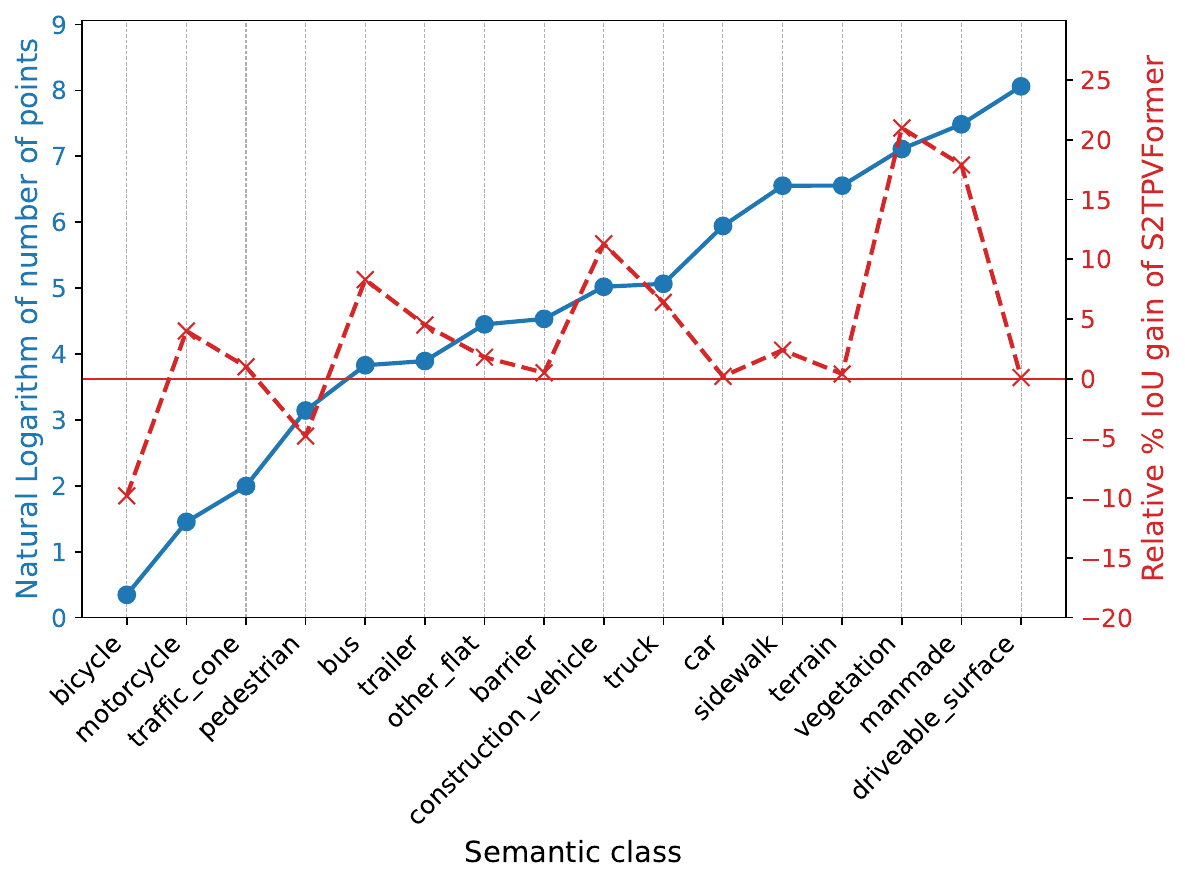}
    \caption{This figure presents a dual-axis representation, where the blue axis and its corresponding graph show the distribution of the natural logarithm of the \textbf{number of per-class ground truth points} in the training dataset. Conversely, the red axis and its graph show the \textbf{per-class IoU gain} achieved by S2TPVFormer in comparison to TPVFormer.}
    \label{fig:points_deltaGain_v_class}
\end{figure}

Conversely, it could also be argued that the observed phenomenon may be partially attributed to the variance in the volume of ground truth data available for each class within the training dataset. In an effort to explore this hypothesis further, figure \ref{fig:points_deltaGain_v_class} compares the number of ground truth points against the comparative performance gain of S2TPVFormer-U (base) relative to TPVFormer, distributed among the sixteen semantic classes. 

Despite bicycles having the fewest ground truth points and pedestrians having the fourth-fewest, TPVFormer surpasses S2TPVFormer in both classes. Conversely, S2TPVFormer excels over TPVFormer in classes like motorcycles and traffic cones, which have fewer ground truth points than the pedestrian class. This indicates that the performance improvement of S2TPVFormer does not directly correlate with the number of ground truth points. This leaves us with our initial hypothesis of S2TPVFormer appearing to be focusing more on larger static objects than smaller dynamic objects.

We see this as a major limitation of S2TPVFormer and assume the reason for the way we compute the unified vision when taking history frames for temporal fusion. We aim to address this in upcoming work.

This comparison aims to learn whether a correlation exists between the number of training data and the model's predictive efficacy across different object sizes and dynamics.

\subsection{S2TPVFormer-W}

\begin{figure*}[ht!]
    \centering
    \includegraphics[width=0.85\textwidth]{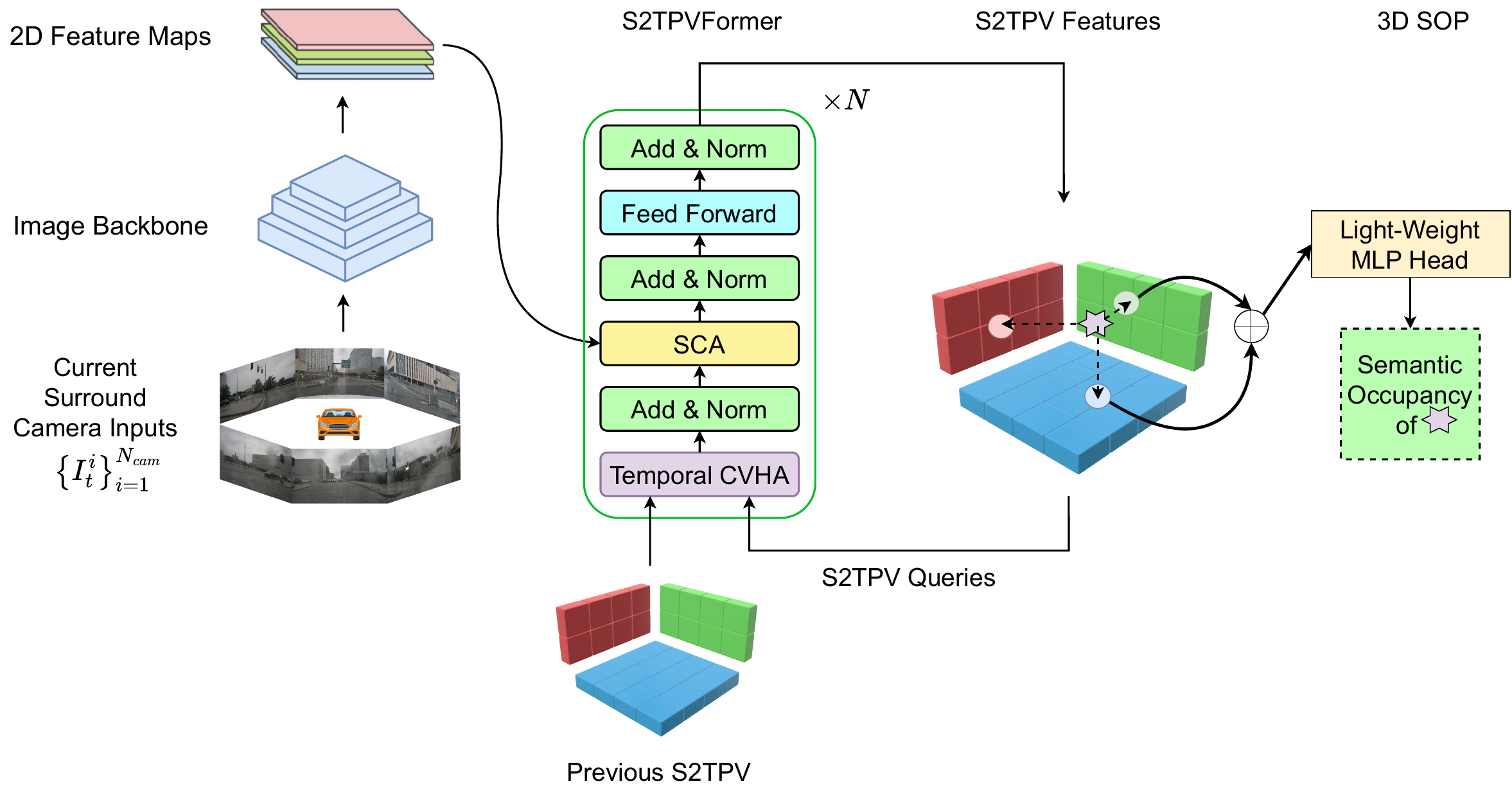}
    \captionsetup{width=\textwidth}
    \caption[width=0.5\textwidth]{The 3D SOP pipeline for the S2TPVFormer-W architecture. (a) First, we utilize an image backbone to extract 2D multi-scale feature maps. We adopt the spatial fusion mechanism of \cite{triperspective} to elevate 2D scene features to 3D TPV latent features. Subsequently, we employ Temporal Cross-View Hybrid Attention (TCVHA) to construct the S2TPV embeddings, enabling the interchange of these spatiotemporal features across the three planes. (b) Inspired by \cite{bevformer}, we concatenate the warped (rotated and aligned with the current ego-space) BEV features from the previous time step with the S2TPV queries at the current layer, achieving temporal fusion.}
    \label{fig:S2TPVFormer_W}
\end{figure*}

In this section, we provide the details of our warp-based temporal fusion architecture, S2TPVFormer-W. Figure \ref{fig:S2TPVFormer_W} shows a block diagram of the architecture. It consists of three major components, which are, the S2TPV Queries, Spatial Fusion via the SCA module, and Temporal Fusion via the Temporal CVHA module. As we discuss in our main paper, the main difference between S2TPVFormer-W and S2TPVFormer-U lies in the view transformation process. We use warp-based temporal fusion for S2TPVFormer-W instead of parallel spatiotemporal fusion, trading off accuracy for compute efficiency.

We present two versions of S2TPVFormer, differentiated by their temporal fusion methods: (a) S2TPVFormer-U, which is thoroughly explored in the paper, and (b) S2TPVFormer-W, which we will examine in this section. S2TPVFormer-W encoder employs warp-based recurrent temporal fusion using TCVHA in place of the self-attention module and Spatial Cross-Attention (SCA) module in place of the Unified Spatiotemporal Module to lift 2D scene features to 3D and fuse them onto S2TPV queries. In S2TPVFormer-W, the BEV features of the S2TPV features computed at the previous time step are preserved and warped according to the ego-motion following \cite{bevformer} to be temporally fused via TCVHA with the current S2TPV queries. Even though the direct temporal fusion via concatenation is done on the BEV plane, TCVHA allows interactions between all three planes in the current queries as well as in the previous BEV features.

\subsection{Depth Consistency}

\begin{figure*}[ht!]
    \centering
    \begin{subfigure}{0.6\textwidth}
        \centering
            \includegraphics[width=\linewidth]{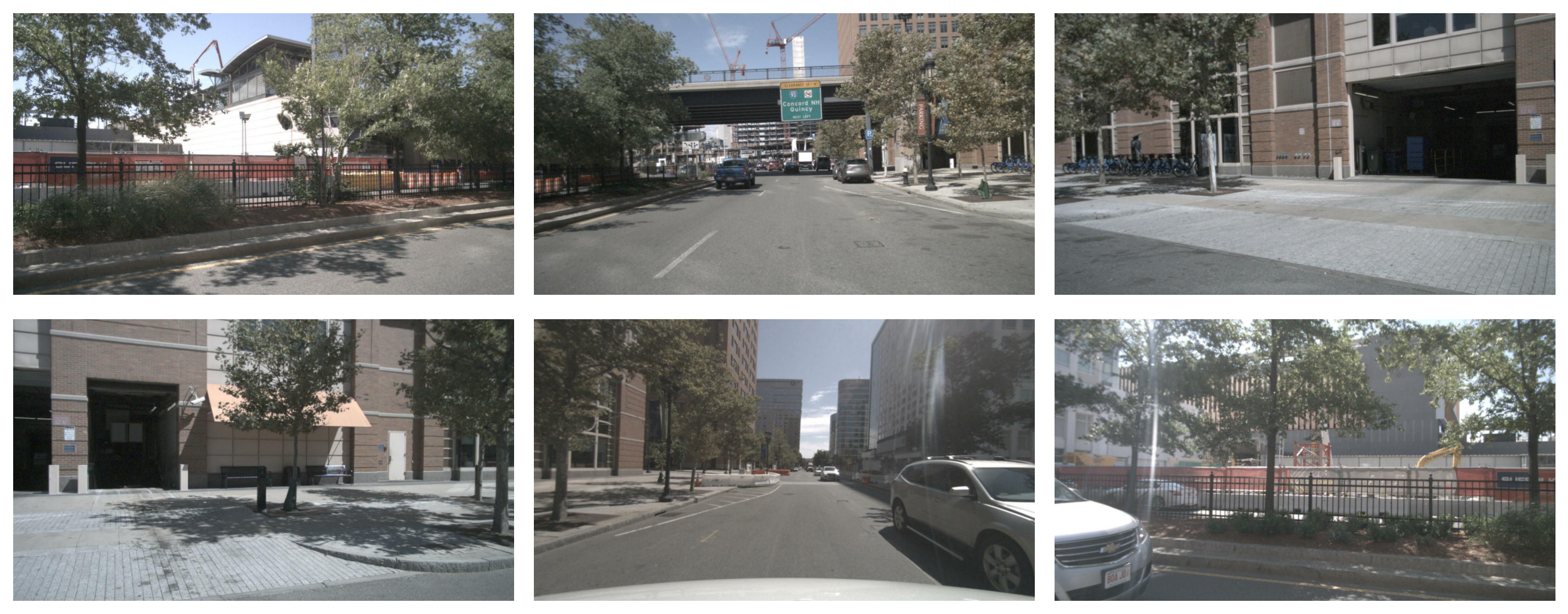}
            \subcaption{Six camera images for a scene.}
            \label{fig:cam_images}
    \end{subfigure} \\
    \begin{subfigure}{0.3\textwidth}
        \centering
            \includegraphics[width=\linewidth]{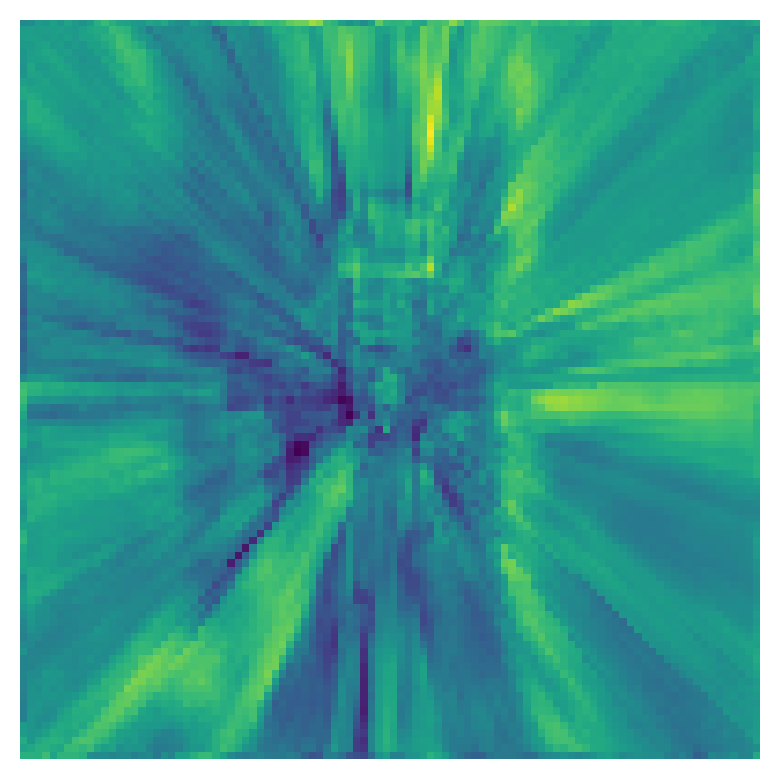}
            \captionsetup{justification=centering, width=.8\linewidth}
            \subcaption{Visualization of TPVFormer BEV plane embeddings}
            \label{fig:embed_vis_tpvformer}
    \end{subfigure}
    \begin{subfigure}{0.3\textwidth}
        \centering
            \includegraphics[width=\linewidth]{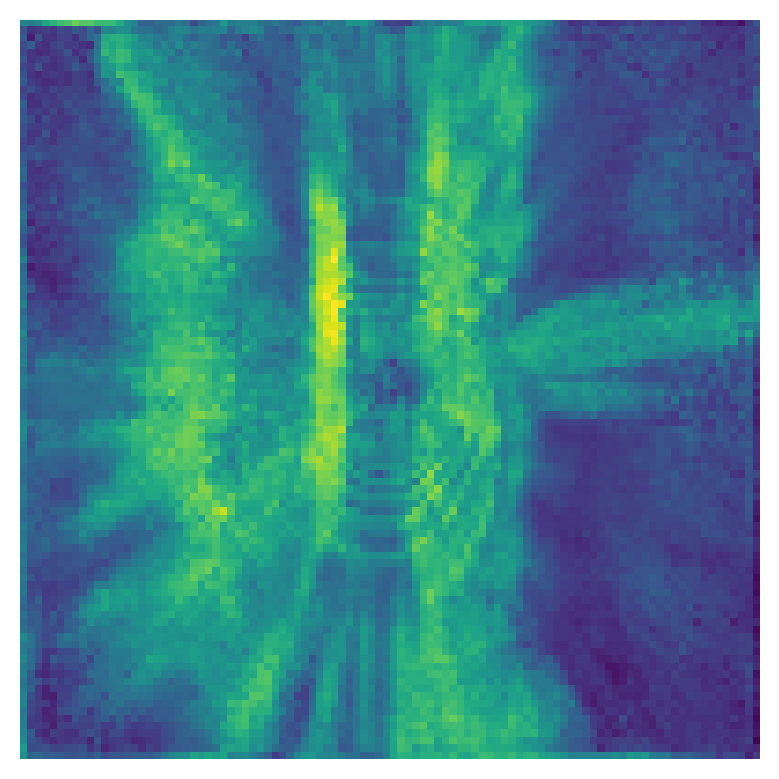}
            \captionsetup{justification=centering, width=.8\linewidth}
            \subcaption{Visualization of S2TPVFormer BEV plane embeddings}
            \label{fig:embed_vis_s2tpvformer}
    \end{subfigure} \\
    \begin{subfigure}{0.3\textwidth}
        \centering
            \includegraphics[width=\linewidth]{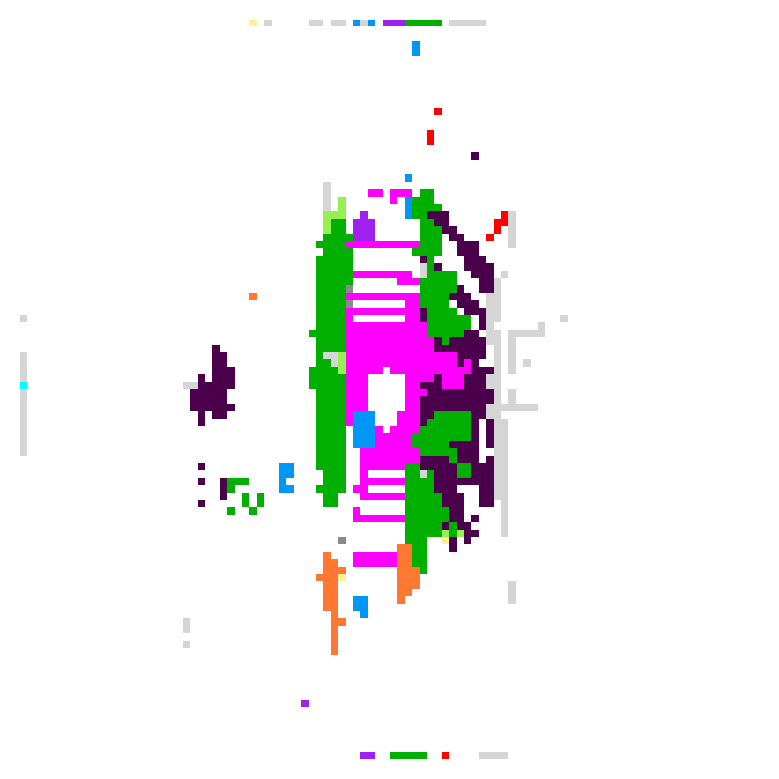}
            \subcaption{TPVFormer SOP Predictions}
            \label{fig:vis_tpvformer}
    \end{subfigure}
    \begin{subfigure}{0.3\textwidth}
        \centering
            \includegraphics[width=\linewidth]{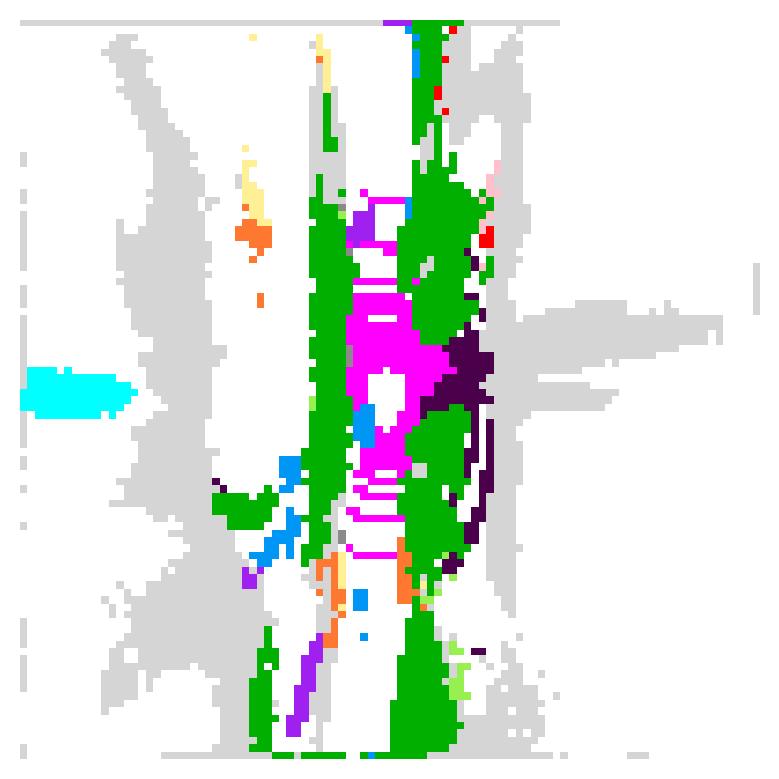}
            \subcaption{S2TPVFormer SOP Predictions}
            \label{fig:vis_ours}
    \end{subfigure} \\
    \begin{subfigure}{0.3\textwidth}
        \centering
            \includegraphics[width=\linewidth]{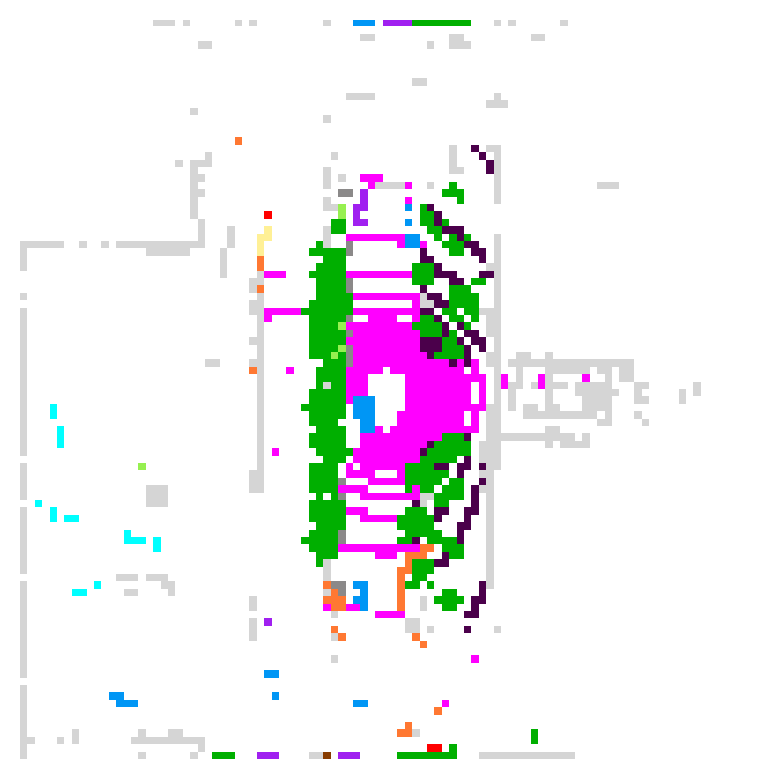}
        \subcaption{LiDAR Ground Truth}
        \label{fig:vis_gt}
    \end{subfigure}
    \caption{\textbf{Comparison of visualizations of the BEV plane embeddings and SOP predictions between TPVFormer \cite{triperspective} and S2TPVFormer-U for the same frame on the nuScenes validation set.} (a) presents the input RGB camera images, (b) and (c) present the BEV plane embeddings from the last encoder layers, (d) and (e) present the SOP predictions from a top-down perspective, with the corresponding LiDAR ground truth presented in (e).}
    \label{fig:embed_vis}
\end{figure*}

\begin{figure*}[ht!]
    \centering
    \begin{subfigure}{0.6\textwidth}
        \centering
            \includegraphics[width=\linewidth]{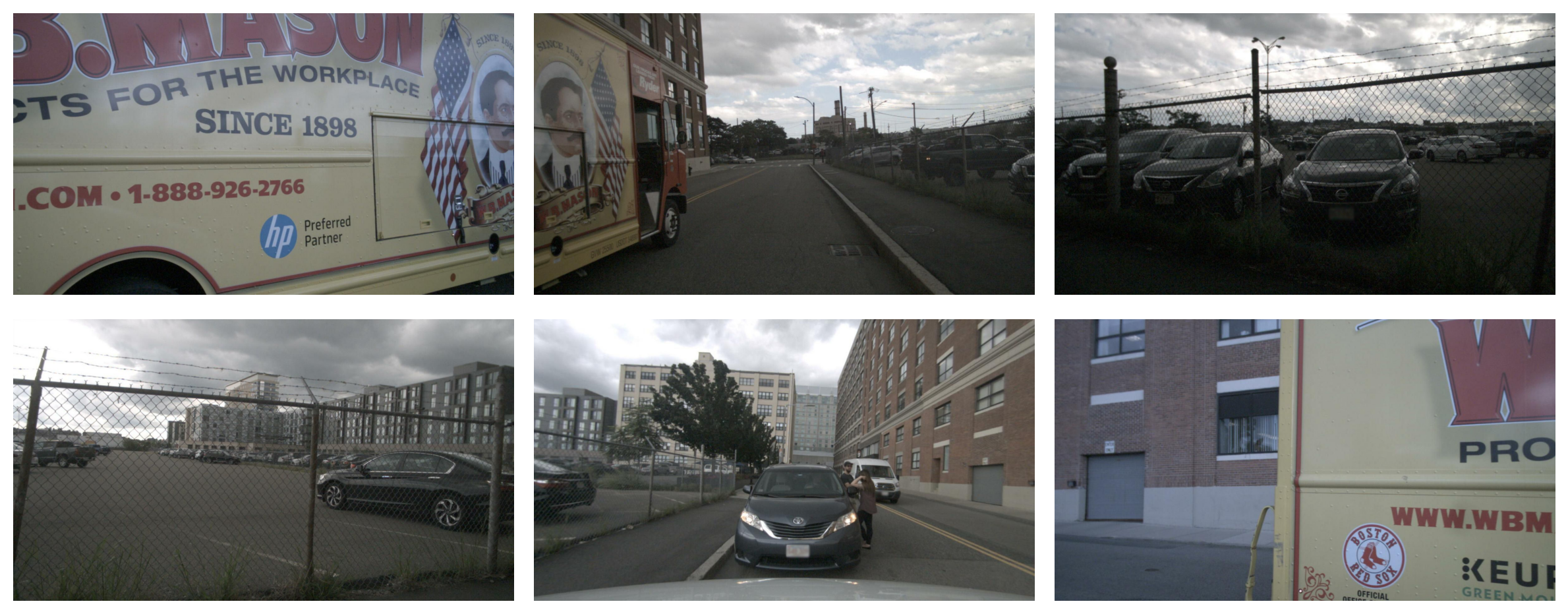}
            \subcaption{Six camera images for a scene.}
            \label{fig:cam_images2}
    \end{subfigure} \\
    \begin{subfigure}{0.3\textwidth}
        \centering
            \includegraphics[width=\linewidth]{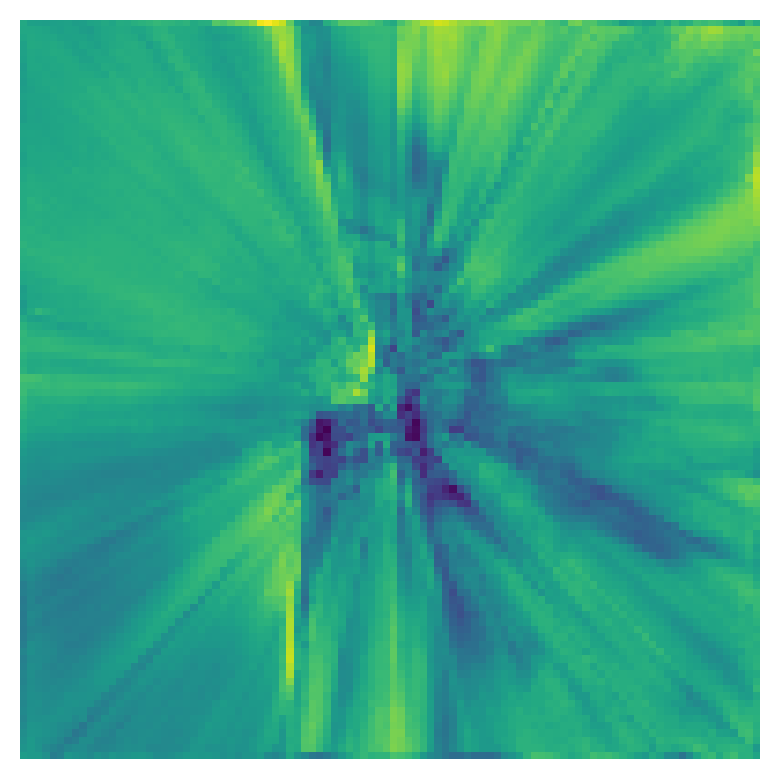}
            \captionsetup{justification=centering, width=.8\linewidth}
            \subcaption{Visualization of TPVFormer BEV plane embeddings}
            \label{fig:embed_vis_tpvformer2}
    \end{subfigure}
    \begin{subfigure}{0.3\textwidth}
        \centering
            \includegraphics[width=\linewidth]{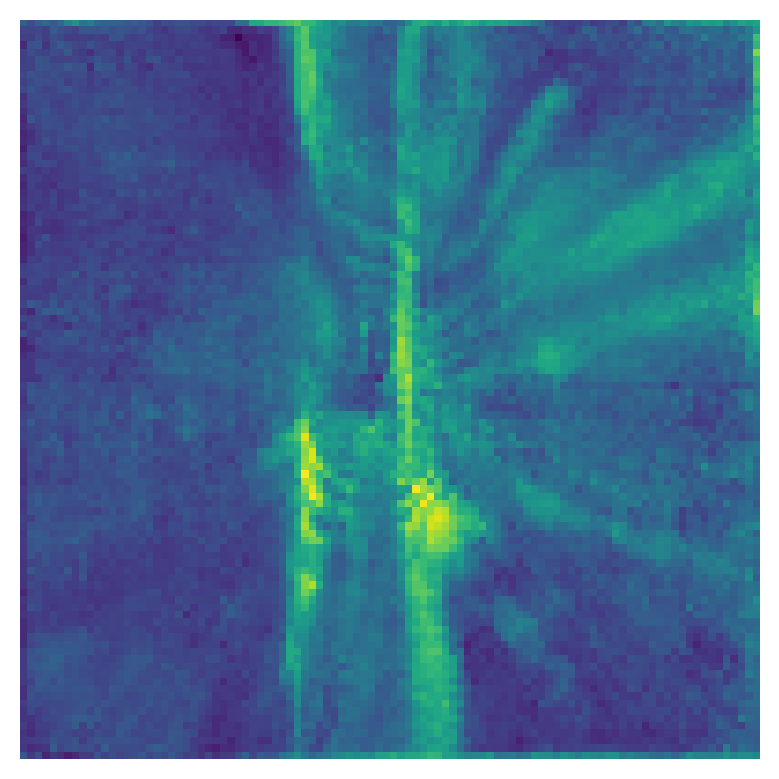}
            \captionsetup{justification=centering, width=.8\linewidth}
            \subcaption{Visualization of S2TPVFormer BEV plane embeddings}
            \label{fig:embed_vis_s2tpvformer2}
    \end{subfigure} \\
    \begin{subfigure}{0.3\textwidth}
        \centering
            \includegraphics[width=\linewidth]{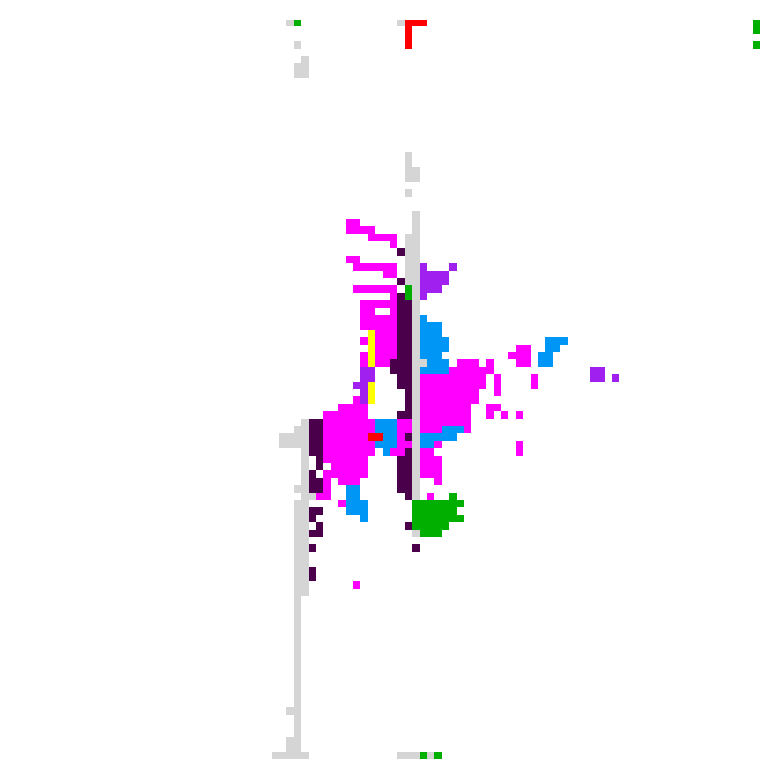}
            \subcaption{TPVFormer SOP Predictions}
            \label{fig:vis_tpvformer2}
    \end{subfigure}
    \begin{subfigure}{0.3\textwidth}
        \centering
            \includegraphics[width=\linewidth]{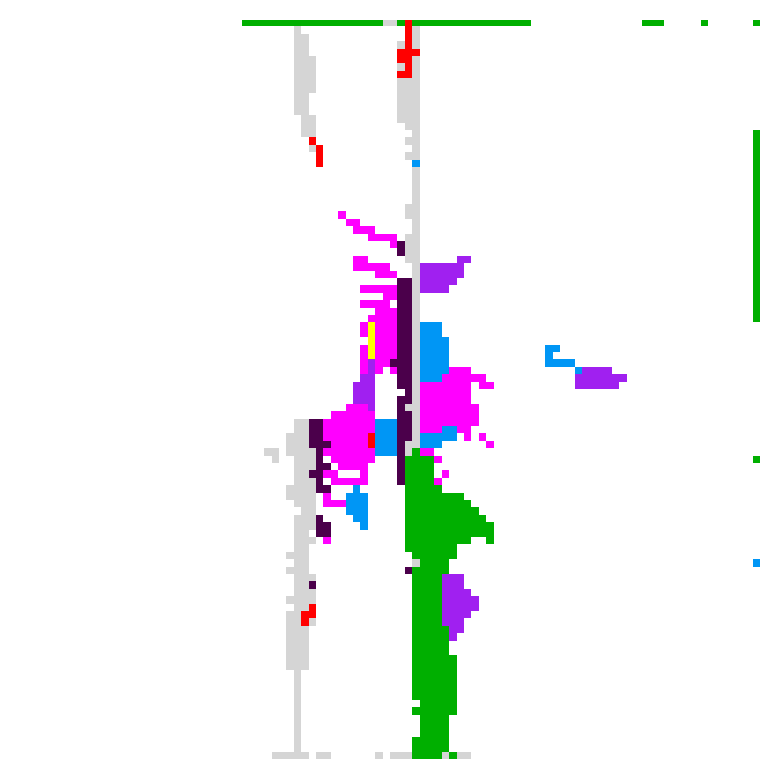}
            \subcaption{S2TPVFormer SOP Predictions}
            \label{fig:vis_ours2}
    \end{subfigure} \\
    \begin{subfigure}{0.3\textwidth}
        \centering
            \includegraphics[width=\linewidth]{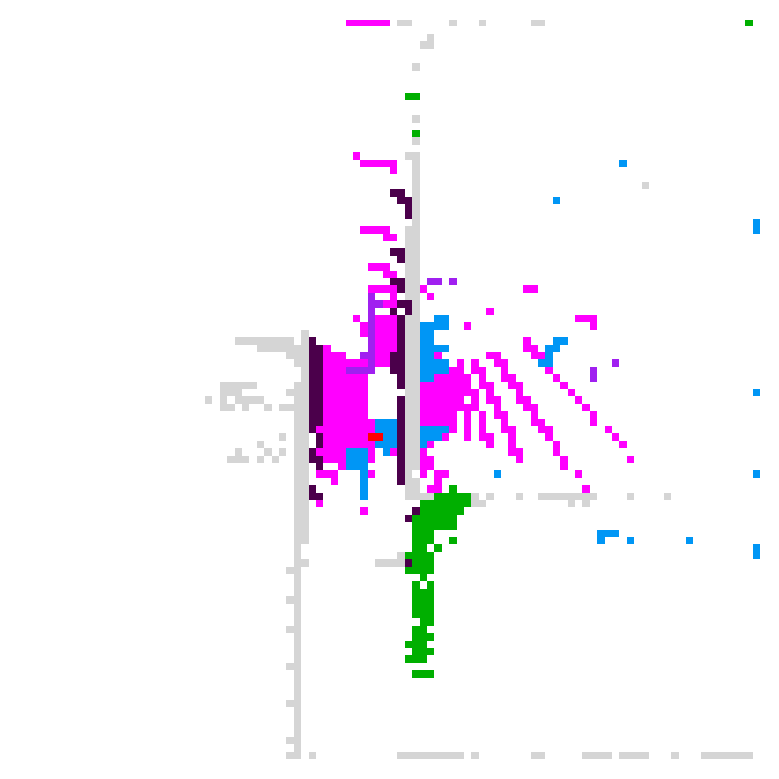}
        \subcaption{LiDAR Ground Truth}
        \label{fig:vis_gt2}
    \end{subfigure}
    \caption{Same set of visualizations shown in figure \ref{fig:embed_vis} for a different scene where the left view of the ego vehicle is occluded by a bus.}
    \label{fig:embed_vis_2}
\end{figure*}

In this section, we analyze the depth awareness of S2TPVFormer-U in comparison to the baseline TPVFormer. As illustrated in figure \ref{fig:vis_tpvformer}, TPVFormer's Semantic Occupancy Prediction results contain numerous false positives (FP) when compared to the LiDAR ground truth shown in figure \ref{fig:vis_gt}. Additionally, the visualization of BEV plane embeddings from the final encoder layers of TPVFormer and S2TPVFormer-U in figure \ref{fig:embed_vis_tpvformer} and figure \ref{fig:embed_vis_s2tpvformer} indicates that ray-shaped features are more distinct in TPVFormer. According to FB-BEV \cite{li2023fbbev}, these ray-shaped features in BEVFormer \cite{bevformer} are due to the lack of depth information during the view transformation process. Specifically, during query-based view transformation, which typically occurs before spatial fusion, 3D points in the ego-space are projected onto the 2D camera images, resulting in a 2D point on an image corresponding to a ray of points in 3D space. This leads to semantic ambiguity for distant objects along the longitudinal direction without depth information, causing TPVFormer to predict false positives along these rays. Conversely, as demonstrated in figure \ref{fig:vis_ours}, S2TPVFormer's predictions show enhanced depth consistency compared to TPVFormer. Further evidence is provided in figure \ref{fig:embed_vis_2}, where TPVFormer's predictions (see figure \ref{fig:vis_tpvformer2}) in heavily occluded areas exhibit ray-shaped artifacts. In contrast, S2TPVFormer's predictions (see figure \ref{fig:vis_ours2}) maintain better depth alignment, resulting in fewer ray-shaped features. We attribute this improvement to the enhanced view transformation facilitated by the unified spatiotemporal fusion module. Consequently, this depth consistency leads to relatively denser semantic occupancy predictions.

\end{document}